# WeedNet: A Foundation Model-Based Global-to-Local AI Approach for Real-Time Weed Species Identification and Classification

Yanben Shen[1]†, Timilehin T. Ayanlade[2]†, Venkata Naresh Boddepalli[1], Mojdeh Saadati[3], Ashlyn Rairdin[1], Zi K. Deng[4], Muhammad Arbab Arshad[3], Shivani Chiranjeevi[2], Aditya Balu[2], Daren Mueller[5], Asheesh K Singh[1], Wesley Everman[1], Nirav Merchant[4], Baskar Ganapathysubramanian[2], Meaghan Anderson[1], Soumik Sarkar[2]*, and Arti Singh[1]*

[1]Department of Agronomy, Iowa State University, Ames, USA.
[2]Department of Mechanical Engineering, Iowa State University, Ames, USA.
[3]Department of Computer Science, Iowa State University, Ames, USA.
[4]Data Science Institute, University of Arizona, Tucson, USA
[5]Department of Plant Pathology, Entomology and Microbiology, Iowa State University, Ames, USA
*Address correspondence to: arti@iastate.edu, soumiks@iastate.edu
†These authors contributed equally to this work.

## Abstract

Early weed identification is crucial for effective management and control, and researchers, agronomists, and technology developers are increasingly interested in automating this process using computer vision and artificial intelligence; however, limited expert-verified data and variable morphological features have hindered the development of AI-based weed identification models. To address these issues, we present WeedNet, a global-scale weed identification model that can recognize an extensive set of weed species. WeedNet is an end-to-end real-time weed identification pipeline that uses self-supervised learning, fine-tuning, and enhanced trustworthiness strategies. WeedNet achieved 91.02% accuracy across 1,593 weed species, with 41% of species achieving 100% accuracy. Using a fine-tuning approach, the local Iowa WeedNet model achieved 97.38% overall accuracy for 84 Iowa weeds. Testing across intra-species dissimilarity and inter-species similarity suggests that diversity in the images collected, spanning all the growth stages and distinguishable plant characteristics, is crucial in driving model performance. The global WeedNet model serves as a foundation, and the global-to-local strategy enables targeted fine-tuning to improve performance in regional weed communities. Additional validation of drone- and ground-rover-based images highlights WeedNet's potential for integration into robotic platforms. Furthermore, integration with artificial intelligence for conversational use provides intelligent agricultural and ecological conservation consulting tools for farmers, researchers, and government agencies across diverse landscapes.

# 1 Introduction

Weeds adapt to natural and human-modified environments, competing with cultivated crops for essential resources such as water, nutrients, light, and space, which significantly affects crop yields[1]. In the United States and Canada, competition from weeds caused an average loss of $26.7 billion in maize (*Zea mays* L.) from 2007 to 2013[2], while wheat (*Triticum aestivum* L.) caused losses averaging $2.19 billion for winter wheat and $1.14 billion for spring wheat during the same period[3]. Dry bean (*Phaseolus vulgaris* L.) faced average losses of $622.2 million between 2007 and 2016 due to weed infestations[4], demonstrating the extensive economic impact of weeds on crop productivity. The National Invasive Species Information Center (NISIC) lists approximately 60 terrestrial invasive plant species, including trees, shrubs, vines, grasses, and herbaceous plants[5]. These species cause significant ecological damage by altering habitats and disrupting native ecosystems (Supplementary Note 1).

In production agriculture, where a herbicide-tolerant crop is grown, a blanket spray with an appropriate herbicide can somewhat reduce the need for detailed weed scouting; however, using it throughout the season is undesirable, as it promotes the spread of herbicide-tolerant weeds and creates long-term production challenges[6]. Herbicide resistance is common, with an average of five new cases emerging annually between 1990 and 2015[7]. For example, Palmer amaranth and common waterhemp (*Amaranthus tuberculatus*), two related species of the Amaranth family, show different herbicide resistance profiles, highlighting the need for accurate identification[8,9].

Correct identification at an early growth stage is crucial for implementing targeted control measures, optimizing weed management costs, and preventing the spread of weed populations[10,11]. Integrated weed management relies on weed identification to establish a multifaceted management program that integrates chemical control methods with cultural, mechanical, and/or biological methods[12]. Invasive weed species pose a serious threat not only to agriculture but also to natural ecosystems. However, the ecological impact of these invasive species is often recognized only after they become widespread, making early detection and classification essential for effective control[5,13].

In practice, farmers are often familiar with the three to four dominant weed species in their fields by observing weed patches during routine field scouting, but a typical agricultural field may be infested with up to 20 weed species[14]. One challenge is that the persistent use of a limited number of herbicides in herbicide-tolerant cropping systems, along with other factors, has led to an increase in herbicide-resistant weed species[15] and a decrease in the effectiveness of glyphosate for weed control[16], as weeds have evolved resistance to glyphosate in the United States, Argentina, and Brazil[17]. Meanwhile, weed seeds and pollen can disperse across farms, further challenging the management of herbicide-resistant weeds[18]. Accurate weed-species identification is important for selecting the most effective control strategies. However, detecting new weed infestations often relies on manual scouting by local professionals and extension specialists rather than digital tools[19,20].

Weed management is moving toward automation, site-specific management, and integration with decision support systems[21–23]. Artificial intelligence (AI) is rapidly emerging as a transformative technology that helps farmers and scouts identify and mitigate plant diseases [24–26], insects[27,28], and weeds[29,30]in a timely manner. Image-based phenotyping has significantly advanced weed detection through non-destructive high-throughput methods for accurate plant classification and spatial distribution[31]. Drones and ground robots equipped with RGB, multispectral, and hyperspectral cameras automatically capture high-resolution field images, supporting the development of algorithms for automatic weed detection and mapping[32,33], diseases[24–26], and insects[27,28]. Remote sensing technologies integrated with AI provide scalable, precise solutions for weed management in precision agriculture, enabling real-time, species-level weed identification. A key challenge is developing AI models that can distinguish subtle differences between crops, weeds, and look-alike species in field conditions, which the PlantCLEF challenge addresses[34]. This global competition uses a dataset of 80,000 plant species to address multiple classes, class imbalance, variable image quality, and noisy field backgrounds. For example, PlantCLEF's 2023 dataset included more than 1.2 million annotated images to train convolutional neural networks (CNN) for fine-grained identification. These advances improve the accuracy of AI-driven weed detection tools while minimizing false positives, bridging the gap between scalable data collection and actionable weed management strategies[35].

Several challenges remain in automated weed identification in crop production. These include (a) limitations in data quality and quantity, which affect species coverage, (b) the need for model development to handle the wide diversity and complexity of weeds, (c) ensuring consistent performance across different species and growth stages, and (d) deploying AI models on autonomous platforms and integrating portable AI workstations for in-field applications.

The main challenge in developing robust AI-based deep learning models to identify weeds in agriculture is the limited quantity and quality of image-based datasets. Quantity concerns include the total number of images and whether each class is adequately represented. High-quality data must include diverse, high-resolution images taken under varying conditions, such as different lighting conditions, growth stages, soil types, and moisture levels. Previous studies have reviewed publicly available weed datasets[32,36,37] (Supplementary Table S1). However, many contain images from controlled environments, such as greenhouses, and lack plant phenological variability[38,39]. In addition, these datasets often collect weed images from their local region, reflecting regional biases and making it difficult to compile diverse datasets for comprehensive AI models. Expanding dataset diversity is crucial to improving model generalization and performance. iNaturalist is one of the largest publicly accessible biodiversity image repositories available for training AI models[40–42], offering research-grade photos with species-level identification[43]. Previous studies have compared research-grade iNaturalist observations with digitized herbarium specimens and reported similarly low misidentification rates between the two sources, supporting the reliability of iNaturalist data for scientific use[44]. However, crowd-sourced biodiversity platforms can include label noise, and we do not assume iNaturalist labels are perfect. Self-supervised learning (SSL) during foundation pretraining can mitigate label concerns. SSL learns visual features directly from images without species labels, making it naturally robust to imperfect annotations and particularly effective at large scale[45,46]. A previous study from our group shows that combining SSL and fine-tuning is not sensitive to label noise[28]. In addition, class imbalance is a concern, as iNaturalist data often underrepresent rare or difficult-to-identify species and overrepresent common ones due to observer preferences and accessibility.

Selecting the appropriate deep learning model for weed identification requires balancing model complexity, computational resources, data availability, and the need for high accuracy in real-world agricultural applications. Early attempts used handcrafted features and traditional machine learning (ML) techniques, such as support vector machines (SVM) and random forests (RF), to differentiate crops from weeds based on color, shape, and texture descriptors[47–49]. However, these models struggled to handle variations in weed morphology and complex environmental conditions, limiting their applicability. Recent studies have used deep neural network architectures, such as CNNs, to improve accuracy under various field conditions. CNN pre-trained architectures, such as ResNet-50, EfficientNet, and DenseNet, have demonstrated success in multiclass weed identification in cotton fields[50]. Researchers have also developed graph-based deep learning models for complex conditions[51]. CNN-based models have been used to classify multiple weed species under various weather and sunlight conditions[52,53]. Semi-supervised methods, such as semi-supervised generative adversarial networks (SGAN), have emerged to address the challenge of obtaining labeled data, achieving high accuracy with minimal labeled data[54]. SemiWeedNet integrates labeled and unlabeled images for weed segmentation and identification, improving model robustness[55]. Despite these advances, traditional deep learning models, such as CNNs, struggle to generalize and scale because training datasets lack sufficient diversity and variability[32,56]. Recent models, such as transformers built around attention mechanisms, including Vision Transformers (ViT), offer better scalability, precision, and efficiency for large-scale image classification tasks than CNNs[57,58].

Weed recognition presents unique challenges, including immense species diversity, wide variation in growth stages, and complex, cluttered backgrounds common in agricultural environments. Unlike CNNs, which analyze images through local receptive fields and fixed convolutional filters, ViTs utilize self-attention mechanisms that process entire images holistically, capturing long-range dependencies and context essential for distinguishing visually similar weeds and deciphering fine-grained morphological variation[59]. This attention-based approach helps ViTs remain robust to shifting scales, rotation, and occlusion in field data while making them more responsive to subtle cues that distinguish closely related species. ViTs scale more efficiently with larger datasets and models, enabling dramatic accuracy gains as image collections grow. This is especially critical for weed identification, where databases often reach millions of images spanning over a thousand classes. Recent work consistently shows that ViT architecture outperforms CNNs on heterogeneous plant disease and weed datasets[60,61].

However, a significant challenge arises when weed identification models provide inaccurate predictions with high confidence, affecting user trust and system reliability[62,63]. Failing to warn users about the model's uncertainty when predicting look-alike weed species can erode confidence in AI applications. Data bias from non-representative training datasets can lead to unreliable predictions. The lack of explainability in complex black box models makes it difficult for users to trust their output, especially when errors occur[64].

The performance of weed identification models in real-world scenarios depends on their ability to accurately detect various weed species under variable field conditions, including common noxious, invasive, and look-alike species. Intraspecies dissimilarity refers to individual differences within the same species[34] (Figure 1a). Common lambsquarters, for example, show considerable intraspecific variation in leaf shape and size, influenced by environmental factors such as soil nutrients and water availability[65]. Horseweed also exhibits distinct visual differences between its rosette and flowering stages, which complicate identification[66–68]. The examples illustrate several plant characteristics, including leaf shape, leaf margin, and changes in plant morphology from seedling to fruiting stage. These variations within a single species across different growth stages highlight the challenge of models in consistently identifying species. In contrast, inter-species similarity involves similarities between different species, including shared physical characteristics and ecological niches[69]. Interspecies similarity occurs when two or more weed species resemble each other in leaf shape, growth habit, flower structure, or seeds[70] (Figure 1b). For example, amaranth species such as redroot pigweed (*Amaranthus retroflexus*), Palmer amaranth, and common waterhemp often co-exist in disturbed habitats, making them difficult to differentiate. Poisonous and non-poisonous species within the Apiaceae family (also known as the carrot family) share traits such as pinnately divided leaves and umbel-shaped flower clusters, complicating identification[71]. Accurate identification is essential because wild parsnip (*Pastinaca sativa*) has phototoxic sap that poses public safety risks. In contrast, caraway (*Carum carvi*) and Golden Alexanders (*Zizia aurea*) have different ecological importance; Golden Alexanders are a beneficial native species, and caraway is a useful culinary herb. Monocot species such as yellow foxtail (*Setaria pumila*), green foxtail (*Setaria viridis*), and giant foxtail (*Setaria faberi*) require close examination of morphological traits. Key identification features include the presence and distribution of hairs in leaf blades, the color and shape of the bristly seed heads, and the structure of the ligules and auricles for accurate identification[72]. Despite differences in toxicity or ecological function, these species share similar phenotypic characteristics, which complicates accurate field identification.

Deploying AI models to detect weeds poses significant computational challenges, as complex deep learning models require substantial computational resources[73]. An ongoing research challenge is balancing algorithm complexity with processing capacity for real-time field applications and optimizing AI models for efficient processing on edge devices[74]. Integrating AI-based weed detection models into edge devices such as smartphones, rovers, and drones requires seamless integration for automated identification and real-time decision-making.

To address the above challenges in data quantity and quality, model generalization and performance, and deployment, this study presents WeedNet. This foundation model (global model) is trained on a large-scale dataset derived from iNaturalist, capturing broad variability across weed species and environmental conditions. However, the iNaturalist dataset has several limitations, including imbalanced species representation, varying image resolutions, noisy backgrounds, and a lack of real field weed images. Leveraging self-supervised learning and supervised end-to-end fine-tuning enables the model to learn robust visual features directly from the images themselves, independent of species names or classifications, making this approach particularly powerful for large-scale datasets prone to label noise[28,75]. This model uses self-supervised learning (SSL) with a ViT architecture to learn general visual patterns across a wide range of plant species from various regions, without relying on labeled data[59]. Although the global model captures rich, transferable features for plant identification, it may not be optimized to detect or classify agriculturally relevant weeds in a specific region or cropping system. To improve relevance and accuracy in regional applications, the global model is fine-tuned using smaller, expert-validated local datasets[28,76]. This transfer learning approach leverages the generalizability of large models while requiring less data and computation, thereby reducing overfitting on limited datasets[77]. This hybrid approach reliably identifies key weeds in targeted agricultural settings while reducing the need for extensive local annotation. By integrating citizen science data with expert labels, WeedNet ensures reliable real-world performance. Consequently, the global foundation model captures transferable visual representations from large-scale, multi-regional weed imagery, while the local fine-tuning step efficiently adapts these representations to region-specific weed communities with limited expert-validated data. This transfer learning framework reduces the need for extensive local annotation and demonstrates the global model's universality, serving as a base model and enabling scalable weed identification across regions worldwide.

# 2 Results

## 2.1 Global Citizen Science Dataset

The WeedNet dataset contains 1,593 weed species and 813 genera, with a total of ~ 14M iNaturalist images, spanning a wide diversity of weed species across classes, orders, and families. The species and taxonomy information are presented in Supplementary Note 2. Common yarrow (*Achillea millefolium*) has the most training images (150,154), while common salt tree (*Caragana halodendron*) has the fewest (182). The median number of training images per species is 3,347, and only 11% of the species have fewer than 500 images. This large, high-quality, diverse dataset enables the deep learning model to capture intra- and inter-species variability, improving its ability to make accurate identifications across fields.

## 2.2 Global WeedNet Model Development

The training campaign was conducted on a large out-of-domain dataset, i.e., a non-weed dataset (3.5 billion Instagram hashtag images), followed by a second stage of SSL pre-training on the in-domain, i.e., weed dataset (iNaturalist), and concluded with supervised fine-tuning on labeled data (iNaturalist + list of all other datasets used here) (Table S5). This model, known as WeedNet, achieved a classification accuracy of 91.02%. The accuracy histogram for all 1,593 weed species shows that 41% of species have 100% accuracy, with a wide range of training images (188 to 15,054). Moreover, 89.7% of species have test accuracy above 80% (Figure 2a). The regression (Slope=0.00223, $R^2 = 0.051$) indicates that increasing training images by 1,000 improves accuracy by only 0.22%, suggesting that the model's performance is largely independent of class size. The intercept indicates that the baseline accuracy (89.0%) applies to species with very few training images, confirming that although some species have far more training samples, model generalization is not dominated by data volume (Figure 2b).

The classification performance analysis compared genus- and species-level results and indicates that misclassifications resulted from intra-genus look-alike confusion and random errors (Table 1). The difference between genus- and species-level performance quantifies how much total prediction uncertainty comes from intra-genus confusion rather than random errors across unrelated species. A 5% accuracy difference indicates that most misclassifications occurred within the intra-genus confusion (e.g., *Amaranthus retroflexus* vs. *A. tuberculatus*) rather than the random errors between unrelated weed species.

Although data imbalance was not the primary determinant of model accuracy, it interacted with intra-genus confusion to cause particularly low performance in a small subset of species. The 1,593-species model showed that 16 species had accuracy below 20% (Supplementary Table S2). Among these sixteen species, most of the misidentifications occurred with other species of the same taxonomic genus (Supplementary Note 3). Moreover, many misclassified images were assigned to species with substantially larger training sets.

To diagnose the reasons for classification differences within the genus, we conducted additional analyses of intra-genus confusion and data imbalance using eight closely related *Amaranthus* taxa. To investigate performance differences due to intra-genus confusion, we clustered latent representations extracted from a Vision Transformer backbone using k-means and used within-cluster variance as a proxy for visual diversity (Supplementary Note 4). Supplementary

Figure **S1** summarizes how within-species diversity changes as the number of k-means clusters increases within the *Amaranthus* genus. At k = 1, *A. hybridus* exhibited the highest normalized within-cluster sum of squares (WCSS), whereas A. retroflexus exhibited the lowest, suggesting that *A. hybridus* expresses broader visual variability in the dataset. In contrast, *A. retroflexus* implies uniform plant appearance across images, reducing within-class dispersion and facilitating discrimination. A distinct pattern was observed for *A. tuberculatus*. While it showed relatively high variance at k = 1, its normalized WCSS declined sharply with increasing k and became among the lowest at large k. This trend suggests that variability within *A. tuberculatus* is not purely diffuse but is instead organized into separable archetypes. When sufficient archetypes are available, and within-cluster variance is small, images of A. tuberculatus can be represented by multiple stable archetypes, which align with its high classification accuracy. To further assess how sampling interacts with visual variability, we compared classification performance under balanced and naturally imbalanced training regimes (Supplementary Table S3). Balanced sampling improved performance for several species, whereas imbalanced training favored certain taxa but degraded accuracy for visually heterogeneous species such as *A. hybridus* and *A. powellii* (Supplementary Note 5).

The training sample size interacted with within-species variance to shape accuracy outcomes. *A. retroflexus* and *A. tuberculatus* species improved with more images, likely because extra data covers each archetype cluster well. In contrast, species with high dispersion, such as *A. hybridus* and *A. powellii*, showed low accuracy under imbalance, consistent with incomplete sampling of within-species variation and increased confusion among visually similar *Amaranthus* species. Together, these results support the conclusion that intra-genus confusion reflects the joint effects of (i) within-species heterogeneity and (ii) insufficient sampling across the various visual patterns present within a species.

## 2.3 Region-Specific Weed Identification through Global-to-Local Fine-Tuning

The Iowa local model resulted in a subset of 84 weed species and achieved an overall accuracy of 97.38%, with most species exceeding 90% per-class accuracy (Figure 2c). However, eleven species exhibited lower accuracies (Figure 2). To address this performance gap, we integrated expert-verified samples from various datasets [50,53,78,79] specifically for low-performing species and fine-tuned the existing model (Figure 2d). The refined WeedNet local model (fine-tuned again), trained with expert-validated data, had an improved accuracy of 97.68% (Figure 2e and Supplementary Table S5). We used a paired t-test to analyze classification accuracy before and after adding 3,017 curated images. Across these 11 low-performing species, accuracy increased by an average of 1.82%, although this improvement was not statistically significant (df = 10, $p = 0.104$). The effect size was small (Cohen's d = 0.42), indicating modest but consistent gains across species. In contrast, the remaining 73 species without additional training images showed no meaningful change in accuracy (mean improvement = 0.07%, n = 73, p = 0.708). These results suggest that incorporating curated expert images can enhance model performance for certain species, but the magnitude of improvement depends on species-specific image variability and the quantity and diversity of new data added. This finding also aligns with our broader observations that accuracy is not driven solely by training image volume; instead, images that capture distinctive, species-specific morphological variability contribute more to model performance.

## 2.4 Performance of WeedNet on in-field and external weed datasets

To substantiate our claim of early high-precision recognition under realistic constraints, we evaluated WeedNet across multiple independent, expert-annotated agricultural weed datasets that differ substantially from iNaturalist imagery in image-capture conditions (Figure 3). Across nine external datasets, the foundation model achieved top-1 accuracies ranging from 25.71% to 72.33% in the zero-shot setting, reflecting the expected domain shift between open-access, heterogeneous imagery and in-field agricultural conditions. Notably, we trained on the 2023 in-field images, while the 2025 images served as an independent test set.

Despite the limited availability of target-domain images, fine-tuning resulted in marked and consistent improvements in identification accuracy (Figure 3b). The largest gains were observed in datasets emphasizing early growth stages and subtle morphological traits. For example, accuracy on PlantSeedlingV2 increased from 25.71% in the zero-shot setting to 86.25% with k = 10 and 91.25% with k = 20. These findings support our central premise that fine-tuning WeedNet enables rapid adaptation for high-precision recognition of early-stage, difficult-to-identify weeds.

## 2.5 Out-of-Distribution Detection Analysis

We evaluated the trained weed model's EBM performance using the same procedure described previously[80,81]. We evaluated uncertainty quantification using three performance measures: AUROC (area under the ROC curve), AURP (area under the precision-recall diagram), and FPR95 (false positive rate at 95% true positive). We choose the OOD threshold based on the ROC curve at -8.9349 to maximize separation between ID and OOD data (Figure 4); we also select the hyperparameter T (temperature) via hyperparameter tuning as 0.02. The results show that the OOD classifier achieved an AUROC of 95.84, an AUPR of 94.08, an FPR95 of 0.00397, and an accuracy of 89.60% in detecting out-of-distribution data. Mapping test data into energy can lead to a clearer distinction between ID and OOD data (Figure 4a).

## 2.6 Conformal Prediction Analysis

In this analysis, we set the hyperparameter $\alpha$ to 0.05, which led us to a threshold $\hat{q}$ of 0.903 on the confidence score. Therefore, the algorithm includes a class name in the conformal prediction set if its confidence score softmax value, is greater than $1 - \hat{q}$. The trained conformal prediction algorithm covered the correct label with 0.947 accuracy. We illustrate an example of the performance of our trained conformal prediction in (Figure 4b).

To further evaluate the model in real-world scenarios, we used it to test challenges to model performance due to intra-species dissimilarity and inter-species similarity. Table 2 shows the quantitative results, and Supplementary Figures S2 and S3 show differences across developmental stages and between look-alike species. The intra-species dissimilarity results show that identifying early-stage plants is difficult, and the model's accuracy improves significantly across developmental stages. The model struggles most with identifying look-alike species, particularly grass species.

## 2.7 Intra-species dissimilarity

The model achieved an overall accuracy of 79.2% on the weed identification dataset (Table 2). Across development stages, the plant's early stage consistently shows lower accuracy than the vegetative and reproductive stages (Table 2). The results show that the model performed better in the later growth stages. As illustrated in the intra-species example (Supplementary Figure S2), early-stage plants have not yet developed most phenological traits, showing only cotyledons and a few true leaves. In contrast, plants in the vegetative stage have fully developed leaves and stems, while those in the reproductive stage show distinct reproductive features, such as flowers and fruits, enabling more accurate identification.

A further experiment used CWD30 dataset[82] to thoroughly evaluate model performance across weed growth stages. The time-series analysis highlights developmental effects on recognition accuracy and shows that fine-tuning improves robustness across stages (Supplementary Figure S4). On this dataset, where images differ markedly from the training data, the foundation model showed low accuracy at the early stage and improved performance as plants developed and diagnostic morphological features became more pronounced. This pattern underscores the importance of incorporating early weed images when assessing real-world weed recognition performance. After fine-tuning, the model produced consistently high accuracy across the developmental timeline. Specifically, the fine-tuned model achieved 88.8% accuracy in the early stage, 94.3% in the middle stage, and 85.0% in the late stage, compared with 34.1%, 54.1%, and 50.0% for the foundation model, respectively. These results indicate that fine-tuning can narrow the early-stage performance gap and improve generalization to different imaging conditions.

## 2.8 Inter-species similarity

Model accuracy for the inter-species similarity, or look-alike species, challenge decreased to 60.6% (Table 2). Comparing accuracy, Macro precision, Macro recall, and Macro F1 score between overall look-alike and life-stage classes indicates that the look-alike challenge is harder to distinguish. In addition, the results show that model performance varied across species families (Table 2). The results indicate that the Apiaceae family outperformed both the Amaranthaceae and Poaceae families in accuracy, macro precision, macro recall, and macro F1 score (Table 2). The model can reach relatively high accuracy, even when images focus only on inflorescence features (Supplementary Figure S3). Additionally, 97.4% Macro Precision indicates that the test species list is a comprehensive dataset that covers most look-alike species with fine-grained leaf features and umbrella-like flowers. The MacroPrecision in the amaranth and grass dataset is 57-59%. This indicates that the global dataset did not include other look-alike species among these eight amaranth species and five grass species. This is because many grass species, in addition to foxtails, in the global dataset and amaranth species have leaf morphology similar to other broadleaf species in other families. Integrating these with the analysis of intra-genus confusion suggests that future improvements in weed recognition should prioritize data quality over data quantity. Capturing key distinguishing features of low-performing species and morphologically similar taxa will be more important than simply increasing dataset size. Such informative features enable the model to consistently learn species-specific visual patterns, thereby improving classification robustness.

## 2.9 Weed Identification on Robotic Platforms

The WeedNet model was evaluated in two different settings: (i) the targeted identification of 15 specific weed species throughout the evaluation period and (ii) an untargeted visual survey assessing the presence of various weeds in the field (Figure 5). When WeedNet was applied to images collected from UAVs and ground-based robots, it accurately identified weeds in both targeted and untargeted evaluations during the early-season dates (Aug 17 and Aug 24). Although the current model functions as an image classifier rather than an object detector, its strong performance on small, early-stage weeds demonstrates robustness across imaging platforms and the ability to correctly classify weed species even when plants are small and image resolution is reduced.

As the season progressed, the model's performance diverged between the two evaluation targets. For the 15 targeted weeds, accuracy remained stable because bias weeds were excluded within the bounding boxes, minimizing interference from surrounding plants and allowing the classifier to focus on the intended species (Figure 5). In contrast, accuracy declined in the untargeted setting, particularly on Aug 30 and Sep 16, when weed density and canopy complexity increased. Untargeted images frequently contained multiple weed species within the frame, resulting in occlusion, overlapping foliage, and visual clutter that hindered the classifier's ability to isolate the correct species. By the final evaluation date (16 September), dense canopy cover, mixed vegetation, and shadowing further reduced feature visibility, contributing to misclassification (Figure 5). These findings highlight the need for fine-tuning WeedNet for object-detection workflows and underscore the importance of selecting appropriate imaging times when deploying UAV or ground-robot platforms for WeedNet classification.

# 3 Discussion

**Data Challenges.** This study represents a pioneering effort to use biodiversity science resources to advance AI models for weed classification. The extensive collection of images and metadata collected, particularly from the iNaturalist dataset, marks a significant milestone and provides unparalleled training data for WeedNet. Most species benefit from more than 500 verified training images, contributed by observers who captured various plant phenotypes, colors, viewpoints, backgrounds, light conditions, and growth stages. Previous studies have often focused on only economically significant or regionally dominant weed species[50,79,83,84]. These models, while effective within their limited scope, lack greater applicability. In contrast, WeedNet is designed for deployment across diverse regions and cropping systems due to its global scale and species diversity.

Furthermore, crop/weed classification models typically overlook species-level weed identification, limiting their applicability in precision weed management, where species-specific herbicide recommendations are critical, especially for differentiating herbicide-resistant from susceptible weed populations[85–87]. WeedNet includes a diverse range of globally relevant weed species, encompassing invasive, noxious, poisonous, and conserved categories. The dataset also includes scientific and common names for each species, supporting consistent taxonomy and improving the reliability of species-level identification across diverse geographic and agronomic contexts.

**Model challenges.** The results show that SSL and fine-tuning strategies can effectively leverage citizen science datasets to improve WeedNet training and accurately identify a wide range of weed species. Local models can improve the prediction accuracy for low-performing species by employing a global-to-local fine-tuning strategy. Furthermore, SSL pretraining on a large, diverse dataset, followed by fine-tuning on small, expert-verified subsets, enables WeedNet to adapt into region-specific, high-performing model variants. In weed detection and classification, fine-tuning pre-trained models on any plant dataset has enhanced the accuracy of supervised learning algorithms[88].

Beyond boosting model accuracy, WeedNet's fine-tuning approach addresses a key AI challenge: in many systems, adding new classes requires retraining the entire model, an expensive, resource-intensive process. WeedNet's global-to-local fine-tuning bypasses this by integrating new species by retraining only the local model with targeted images, leaving the large global model untouched. In a case study, we fine-tuned the Iowa model to two newly observed invasive species found in Iowa, common copperleaf and Asian copperleaf, achieving 96.98% accuracy across 86 species, nearly matching the original 97.38% accuracy across 84 species (Supplementary Note 6). Built on a 1,593-species plant-specific database, WeedNet inherently captures diverse botanical features, enabling rapid, low-cost expansion for local weed classification while maintaining state-of-the-art performance.

One notable limitation our analysis reveals is that the combined effect of intra-genus confusion and class imbalance reduced model performance. Although WeedNet performs strongly overall, the quantitative comparison between genus- and species-level predictions shows that most errors arise from misclassification among closely related species rather than random mistakes across unrelated taxa. This is further supported by the 16 lowest-performing species, where every misidentified sample was consistently predicted as a morphologically similar species within the same genus. Class imbalance amplifies this problem: rare species often have far fewer training images than their common or invasive relatives. As a result, underrepresented species are overshadowed by dominant species with tens of thousands of images, which biases the model toward these majority classes during training. However, this pattern does not mean more images lead to higher accuracy for all species; rather, it reflects the model's inability to extract distinguishable species-level features when phenotypes look alike, and training samples are limited. In other domains, class-imbalance learning approaches such as advanced multiclass algorithms[89], transfer learning with resampling[90], or reweighting techniques[91] have been proposed to address this challenge. However, these weed-identification methods require collaboration with weed scientists to define minority classes, since species with few images but distinct features can still achieve high classification accuracy. Encouraging farmers, researchers, and agricultural extension workers to contribute labeled images can improve model performance and build a more comprehensive database. WeedNet can reduce inconsistencies in crowdsourced datasets by integrating expert data with crowdsourced contributions.

Because ensuring WeedNet's reliability and trustworthiness is critical for practical applications, we incorporate uncertainty quantification and out-of-distribution detection mechanisms. These features help identify cases where the model lacks confidence, prompting users to seek expert verification[28,80]. By incorporating confidence scores and visual indicators of uncertainty, users can make more informed decisions and mitigate the risks of misidentification. These features can also help identify new invasive species in a region when the model consistently reports that captured weeds are out of distribution, and they can help farmers investigate further to identify the invasive species correctly.

**Performance challenges.**

Look-alike species are a major challenge for plant classification; species that differ only in a subset of morphological traits are harder for deep learning models to classify. A previous study thoroughly quantified species distances learned by deep learning models and the morphological trait space of both similar and distinct phytoplankton species, including variation across shapes, within genus groups, and multiple image sources of algal strains[92]. The results indicated that deep learning model representations are highly correlated with morphological features, which is consistent with the findings of this study. LifeCLEF 2016[93] highlighted whether to use whole-plant images, single organs, or multiple organs for species identification, and the training dataset included both species and plant-organ labels such as leaf, flower, fruit, bark, stem, and branch. Prior work comparing different plant organs and whole-plant images generally shows that using multiple organs outperforms using a single organ (for example, flowers alone) and also outperforms whole-plant images[94–96].

Rzanny et al. (2022)[97] conducted an experiment on Poaceae (the grass family) to evaluate identification accuracy using different perspectives and plant parts, including the inflorescence, collar region (leaf neck), leaf tip, leaf underside, and leaf nodes. The authors trained separate models for each organ, as well as a fused model combining all organs. Their results showed that images of the inflorescence and the ligule conveyed the most species-discriminative information. The authors concluded that "even species considered very difficult to identify can achieve high accuracy in automatic identification as long as images depicting suitable perspectives are available." In this study, we also show examples of using images of the grass collar region for prediction (Supplementary Note 7). This study emphasizes that future studies can utilize distinguishable morphological features and improve recognition performance for look-alike species.

Intra-species dissimilarity and differences in developmental stages indicate that differentiating plant characteristics drives model accuracy (subsection 2.7). In the early stage and for grass species, fewer characteristic features result in lower accuracy and more misidentification than in other testing categories. Previous studies also report that most misidentification results from similar morphological characteristics, especially during the early stage, when they can be discriminated only by texture and color[98,99].

Nevertheless, image-based identification faces challenges for certain species due to large intraspecific variation, small interspecific differences, limited training images, and variation in image acquisition conditions[100]. To improve identification performance, incorporating multiple types of observational information has shown greater benefits than relying on single image classification alone. Metadata such as date and geographic location have been used as contextual information in multi-input models[41,101]. Location and time information can infer local weather conditions, habitat type, and observer expertise. For weeds, clusters of visually similar weed images collected in close proximity are more likely to belong to the same species, as indicated by the weed distribution study[102]. Geolocation information therefore helps distinguish look-alike species observed across different regions. In addition, using multimodal data has shown significant improvements compared with image-only models[41,101,103]. Furthermore, for look-alike species where image alone does not work, DNA-based assays may provide higher discriminatory power. DNA markers are more costly but promising. A recent study comparing species classification using DNA-based, image-based, and multimodal models showed that image-based models exhibit weaker genus-level discrimination than DNA-based models, whereas images can outperform DNA in reducing intra-genus confusion for some genera. Integrating DNA and image data consistently yields the lowest confusion rates at both inter-genus and intra-genus levels, demonstrating the complementary strengths of the two modalities[104]. Because DNA-based identification requires sufficient genome coverage, species-level variability, and high data quality, this strategy can identify emerging herbicide-resistant weeds that may cause substantial agricultural damage.

Although look-alike species and very early-stage weeds can cause species-level misidentification, this issue is often secondary in field-scale weed management, where the primary objective is effective control rather than exhaustive botanical cataloging. The fundamental premise of linking model confusion with management groups is that weed control decisions are driven by treatment response, not solely by taxonomic precision. If multiple species within a field respond similarly to the same herbicide or mechanical control strategy, a classifier that groups them into a shared management category remains highly effective for operational decision-making.

In practical herbicide applications, farmers commonly use tank-mix herbicide formulations, which combine multiple modes of action to improve control efficacy and robustness[105]. This management-oriented concept has previously been applied to precision spraying systems. For example, Gutjahr et al. grouped weed species into annual broadleaf weeds, grass weeds, and *Galium aparine* in winter wheat, and into annual broadleaf weeds, grass weeds, and perennials in maize, to guide multiple-tank sprayer design and application strategies[106].

While WeedNet supports such management-level grouping, it also provides species-level taxonomy and ecological information through the WeedID application. Our analysis shows that most misclassifications occur among species within the same genus, which often share similar morphology and management responses; nevertheless, the model achieves 95.78% genus-level accuracy. Importantly, WeedNet reliably identifies highly competitive and economically damaging weeds, such as velvetleaf, waterhemp, Palmer amaranth, and giant foxtail. They are characterized by aggressive growth, herbicide resistance, and significant yield impacts[107]. Accurate identification enables farmers to apply targeted herbicide products and management strategies.

**Deployment challenges.** The model tested on images from the robotic platform showed that citizen-sourced images capture weeds under diverse conditions and that the deep learning model can effectively extract distinguishing features and perform well across platforms. When comparing manually captured images with those from robotic platforms, the primary difference lies in image resolution and the amount of plant detail captured. Previous studies have shown that image resolution drives the accuracy of individual weed plant discrimination[108,109]. In particular, the UAV images in this study achieved resolutions of 0.1 to 0.12 cm/pixel, significantly contributing to the model's effectiveness by preserving fine morphological traits. Meanwhile, a subset of citizen images focused only on specific plant characteristics (e.g., leaves, flowers, or stems), whereas the robotic platform typically captured the entire plant. These differences between citizen datasets and robotic applications highlight the importance of comprehensive plant-characteristic coverage and high-resolution image data when training and deploying models for weed discrimination across imaging platforms.

Deploying WeedNet models on smart devices including smartphones, drones, and ground robots brings substantial real-world value but involves several practical challenges that must be addressed for seamless operation. Edge devices typically have limited memory, processing power, and frame-rate capacity, so we need to optimize our ViT model using compression techniques, quantization, and low-parameter architecture choices to maintain fast, resource-efficient inference with minimal accuracy loss[110]. Platforms such as NVIDIA Jetson Orin Nano or Jetson Xavier are commonly used in robotic and agricultural edge systems; however, their significantly lower GPU memory and compute capacity would likely require model compression techniques (e.g., pruning, distillation, or INT8 quantization) to deploy the full model efficiently. Investigating such compression strategies for Jetson-class edge devices is an important direction for future work. Real-time responsiveness is critical, as robotics applications rely on prompt predictions for navigation and control, and any latency from multiple concurrent processes can impact decision-making and field operations. Network reliability is also a concern, since agricultural environments often have weak or unstable connectivity, making cloud inference and updates unpredictable[111]; therefore, prioritize on-device inference and robust local processing. Finally, as deployments are scaled to multiple devices, backend systems must be designed to handle higher loads and maintain low-latency performance for all users[112]. These combined strategies will ensure WeedNet remains efficient, responsive, and reliable across a range of deployments.

For future applications, WeedNet could be extended for multi-object classification and object detection in complex field settings. High weed infestations often result in overlapping plants, making it difficult to distinguish between species in a single image. Furthermore, variations in lighting and shading across the image can pose additional challenges[53]. A future study will pursue a follow-up study utilizing detection transformer models (DETR)[113]. The WeedNet encoder would serve as the backbone feature extractor, and the DETR decoder would then operate on a set of object queries. Fine-tuning this WeedNet-initialized DETR model on UAV and ground-robot images with bounding-box annotations would enable simultaneous localization and species identification of multiple weeds per image, providing a direct pathway toward autonomous, species-specific spraying and robotic weed management.

WeedNet can also expand beyond identification to include segmentation and quantification of weed density estimation. Using the trained classification encoder for a segmentation model and explainable AI, images can be segmented to identify the spatial distribution of weeds without significant annotation cost[30,114]. Farmers can gain insight into weed infestation patterns and plan targeted interventions. Furthermore, density estimation can quantify weed coverage in fields, providing data-driven recommendations for site-specific weed control. These advancements will enhance the model's utility, making it an indispensable tool for precision agriculture and large-scale weed management strategies. Looking ahead, we plan to broaden WeedNet's scope by developing models for plant groups beyond agricultural weeds. This includes creating dedicated local models for plant biodiversity, forest tree species, and prairie plant species[42]. Beyond expanding species coverage, we aim to integrate advanced AI technologies, such as multimodal learning[115] and explainable AI[116,117], to further boost identification accuracy, adaptability, and interpretability.

The study developed a global-scale weed identification model, WeedNet, using a citizen-sourced image repository (iNaturalist) with over 14 million images representing 1,593 weed species across categories such as invasive, poisonous, noxious, and conserved plants. We used an end-to-end WeedNet pipeline, training with a strategy that combines SSL and fine-tuning using advanced architectures, including MAE and ViT, to improve accuracy and robustness. To further improve prediction trustworthiness, we added two wrapper models: OOD detection and conformal prediction. WeedNet achieved a 91.02% accuracy across all 1,593 species and 97.38% accuracy in 84 Midwest US weed species. The model was further evaluated under real-world challenges, such as intra-species dissimilarity and inter-species similarity, showing that distinguishing morphological traits strongly correlates with accurate classification. Furthermore, successfully predicting captured images from UAV and robotic platforms confirms the model's adaptability to varying resolutions and lighting conditions, highlighting its potential for future real-time weed identification. Integrating the WeedNet API with the chatbot API yields the WeedID application, establishing it as a powerful and intelligent decision-support tool for farmers and researchers in modern agriculture.

# 4 Methods

## 4.1 Dataset

We obtained the image dataset from iNaturalist, which consists of approximately 257 million observations spanning five kingdoms, including approximately 172,000 plant species[118]. We carefully curated the image dataset with the list of invasive species profiles (60 species, NISIC, USDA)[119], Midwest Invasive Plant Network (458 plant species in the US and Ontario, Canada)[120] and Bugwood images (1,916 plant species)[121] to filter it down to 1593 species categories of relevant plants, including crop-specific weeds, noxious weeds, invasive weeds, and weed species from around the world (Supplementary Note 2). In addition, the dataset also includes crop species for application purposes (Supplementary Note 2). We excluded species with fewer than 100 research-grade images in the iNaturalist dataset. The final dataset comprises 14 million images, which were verified to ensure accurate species labels. Figure 6 illustrates the WeedNet workflow, including data resources, model development, dataset information, and model deployment on ground rover and drone platforms. We used representative images and the global distribution map from iNaturalist to illustrate example images and image resources collected worldwide[118]. In addition, we collected image metadata and geolocation information for 84 Iowa weed species to map their observed locations across Iowa[118].

Although iNaturalist data may contain label noise and imperfect annotations, self-supervised learning and fine-tuning can partially mitigate these issues. For a very large population, the precision of a proportion estimate depends primarily on the number of independently reviewed samples rather than on the percentage of the total population sampled. To quantify potential label noise within the iNaturalist subset used in this study, we conducted a complementary sanity check. Domain experts independently reviewed a random sample of 20 images per species across 84 Iowa weed species, following the Iowa State University Extension and Outreach Weed Identification Field Guide[122]. Of the inspected images, 51 were mislabeled. The observed error rate was 3.04%, with a 95% confidence interval of +/- 0.82% (2.22%, 3.86)[123,124]. Most misidentifications resulted from confusion among species within the same genus, mixed vegetation within the image frame, incomplete or ambiguous plant structures, or clearly unrelated or non-plant photographs. With a possible error rate of 4%, SSL pretraining learns visual representations entirely from raw image content without using any class labels, making the representations immune to label noise. Extensive machine learning research already shows that models fine-tuned on self-supervised visual backbones retain high generalization performance under label noise levels well beyond 5%-8%[46,125].

We performed data acquisition and transformation using the Large Data Acquisition Workflow Template (LDAWT), a specialized software suite. This suite facilitates downloading and querying the iNaturalist database, generating file manifests based on specified criteria. It also optimizes manifest groupings to improve download throughput and performance. Using a template-based approach, LDAWT transforms and organizes data during the download process, thereby minimizing the subsequent renaming and reorganization necessary for model training. These manifest files are then transmitted to TaskVine, a parallelized, distributed, and fault-resistant workflow manager, enabling concurrent asynchronous downloads from geographically diverse locations, including cloud virtual machines and high-performance computing (HPC) data transfer nodes. LDAWT aggregates the final output in a machine learning-ready format and deposits it into a shared cloud storage repository. LDAWT uses available network bandwidth cooperatively, preventing denial of service to data providers such as iNaturalist while transferring terabytes of data.

## 4.2 Global Model - Pretraining and Fine-tuning Approach

Accurate machine learning models rely on annotated datasets, where each image is labeled with a corresponding category. However, large-scale dataset annotation remains a significant bottleneck, especially when expert verification is required. To address this challenge, we adopt an SSL approach (Figure 6), which lets the model learn meaningful representations without labeled data. This approach is especially robust for the 14 million datasets, as it is inherently resilient to label noise and removes the need to manually check every label[45,46]. This enables effective feature learning that can be later fine-tuned using a smaller labeled dataset, improving model generalization and performance. We used twenty images per species from the iNaturalist data for testing and validation. To comprehensively evaluate our model's accuracy, we also assessed its performance on a diverse collection of public-domain web images and images collected from our field test, representing various stages of weed species development. We collected 6,962 images from April to November 2023 in corn, soybean, mung bean, and field pea fields in Ames, Iowa, USA. The dataset included 18 Iowa weed species at multiple growth stages and was captured using iPhones (Apple, USA) at a resolution of 3024 × 4032 pixels. We collected an additional in-field dataset in 2025 for testing; it covered 15 weed species and included 3,039 individual plants (Figure 3). Notably, we used the 2023 in-field images for training, while the 2025 images served as an independent test set. The publicly available web images for the global weed species were sourced from government agencies, university extension programs, and academic repositories, excluding iNaturalist, to assess identification challenges. We tested several thousand images separately for the four challenges described in the Introduction.

SSL pretraining follows a pretrain-then-finetune paradigm, which has proven successful in various computer vision tasks, including image identification and object detection[28,126,127]. Instead of relying on supervised labels, SSL methods learn general-purpose visual representations by solving a pre-training task on large-scale unlabeled data. Early approaches focused on reconstruction-based objectives[128], alongside other techniques such as contrastive learning[129] and joint embedding methods[130,131]. With the emergence of ViTs[59], reconstruction-based strategies have regained interest due to their simplicity and state-of-the-art performance[132,133]. A previous study compared CNN-based architectures such as ResNet and RegNet for image classification tasks and selected RegNetY32 for InsectNet because of its efficiency and accuracy for insect identification[28]. Following a similar protocol, we trained and compared a CNN-based model (RegNetY32) and a Transformer-based model (Masked Autoencoder, MAE) to evaluate their supervised classification performance on our 1,593-class weed dataset. The Transformer-based approach outperformed the CNN-based model (Supplementary Note 8 and Supplementary Table S4a), consistent with reports of superior Transformer performance in related plant classification tasks[59,134,135]. We also evaluated multiple MAE model sizes and assessed the effect of pretraining dataset size, finding that the 633M-parameter MAE variant, combined with the larger pretraining corpus, performed best (Supplementary Note 8).

In this work, we use MAE[133], an SSL strategy that has demonstrated superior performance in multiple transfer tasks[133,136,137] while remaining computationally efficient (Supplementary Note 9).

Supplementary Figure S5 shows the MAE model architecture and fine-tuning. Our pipeline involved multiple data preparation steps (Supplementary Note 10), followed by two pre-training levels: initial pre-training on a large-scale dataset of 3.5 billion images with Instagram hashtags, using the same methodology described in[127]. Then, 14 million unlabeled weed images followed a subsequent pre-training. We performed end-to-end fine-tuning on 14 million labeled images (Supplementary Table S5). We conducted extensive training experiments on multiple dataset sizes and model sizes to systematically assess the impact of SSL pre-training. We evaluated the fine-tuned models using standard identification metrics (top-1 and top-5 accuracy) to assess the effectiveness of our approach (Supplementary Table S4b and S4c). The core biological and ecological knowledge captured by WeedNet originates from large-scale domain-specific iNaturalist data, with Instagram pre-training used primarily to accelerate and enhance early feature learning and to maintain methodological alignment with recent best practices in the field. This two-stage pretraining strategy leads to a more robust and generalizable classifier[28,138].

## 4.3 Global-to-local Finetuning

We adopt a hierarchical strategy to accurately identify weeds in the local context. This process begins by leveraging the trained WeedNet model (global model) and implementing a global-to-local fine-tuning approach to develop a region-specific weed identification model for the Midwest US (Figure 6 and Supplementary Table S5). This method began with a robust global model trained on 14 million images. Stages 1 and 2 represent the self-supervised pretraining process using ViT-based MAE on large-scale unlabeled datasets: 3.5 billion Instagram images and 14 million iNaturalist plant-related images. Stages 3 and 4 involve supervised fine-tuning using 14 million labeled iNaturalist weed images, while Stage 4 also adds expert-verified datasets, including open-access weed datasets from other regions around the world[50,78,79,84]. Uncertainty quantification features such as conformal prediction and OOD detection (Stage 5) were integrated to improve the trustworthiness of the model's output. Stages 6 and 7 demonstrate local model fine-tuning using 2 million labeled regional weed data. Stage 8 involves evaluating the model on other open-source, multiple expert-curated datasets under k-shot scenarios (using a certain number of images for fine-tuning). Across the pipeline, SSL pretraining combined with fine-tuning strategies trained on citizen and expert data sources developed WeedNet and Global-to-local models. We then fine-tuned these models using an expert-verified dataset (Supplementary Table S5), focusing on the 84 prevalent weed species affecting Midwest corn and soybean cropping systems[122].

## 4.4 Improving Model Trustworthiness

Two wrapper models, namely out-of-distribution (OOD) detection and conformal prediction, are incorporated into the model to enhance trustworthiness. OOD detection helps avoid misclassifying unfamiliar inputs as known classes, which reduces trustworthiness. OOD detection mitigates this by helping models recognize data that deviates from the training distribution, improving reliability[139–143]. The OOD algorithm assigns a confidence score indicating the likelihood that a test sample belongs to the training set; factors such as model architecture and class distribution influence its effectiveness. Saadati et al.[80] recently evaluated these factors and demonstrated that energy-based models (EBMs)[142] outperform alternatives on multiple performance metrics.

In this paper, we selected EBM as the OOD algorithm and wrapped it around the trained classifier, following the same analysis procedure. For data preprocessing for model (EBM) training, we used the test set of an in-distribution weed dataset (ID), selecting five random samples from each of the 1593 species, with a total of 7,965 images. We used three (60%) to train the EBM and reserved two (40%) to test it. As OOD datasets, we used Human Face[144], ImageNet[145], and Insects, Birds, Non-Weed datasets from iNaturalist[146]. Specifically, for each dataset, we used the following (train/test) image counts: ImageNet (973/973), HumanFace (600/300), Insecta (1526/1526), and non-weed (704/528). The non-weed dataset includes images from different parts of the plants, soil, and fungi. To ensure the ID and OOD test and training dataset sizes are nearly identical, we randomly removed a small number of samples from the OOD dataset to match the ID dataset size.

To enhance the trustworthiness and accuracy of the weed detection model's predictions, we incorporated conformal prediction into the model. When the model is uncertain about a prediction, it is more reliable to output a set of potential labels rather than a single one. Conformal prediction[147] supports this by generating sets of labels based on a confidence level $\alpha$, ensuring that the true label is included with high probability.

For our task, we considered the hyperparameter $\alpha = 0.95$. Our approach consisted of the following steps, using the same hyperparameter setting as in paper[81] while conserving the structure of the original algorithm[147]: First, we took eight random samples from each of the $1593$ species from ID test data to create the conformal training set (4 images per class, 50%) and the conformal testing set (4 images per class, 50%). Then, we select a heuristic function, such as the maximum softmax probability, to measure the conformal score for all images in the training set. We calculated the score for each data point in the conformal training set and then sorted all scores. For a given threshold $\alpha$, the score value of $1 - \alpha$ for the dataset is more significant than that score, which is denoted by $\hat{q}$. In the application phase, we compared each class's softmax value with $\hat{q}$, and if it was higher, we included that class in the conformal set.

## 4.5 Model Performance

To evaluate the WeedNet model's performance on intra-species dissimilarity, we selected a subsample of 84 weed species important to Iowa[122]. We used web-sourced images for each selected weed species, capturing early plant stages (seedling to 3 true leaves), vegetative stages, and reproductive stages, to test the model's performance. Because online images for some species across all stages were limited, we standardized the number of images per species based on the least-available stage, ensuring a consistent ratio across the three growth stages. To evaluate the model's performance on inter-species similarity, the web image set included species previously reported as lookalike species in three categories: Grass family (5 species), Apiaceae family (11 species), and Amaranth family (8 species)[148–151] (Supplementary Table S6). We collected images of each species using the same developmental-stage-based strategy described above.

## 4.6 Model Deployment on Autonomous Field Platforms

To evaluate WeedNet's deployment, we used weed images captured with UAVs and ground rovers during field experiments in Ames, Iowa, USA, in 2024. Image collection began three weeks after planting, at the 2- to 3-true-leaf stage, using UAVs and rovers to capture weeds at various growth stages in inter-row areas and field borders. A DJI Inspire 2 drone (*DJI, Shenzhen*) equipped with a Zenmuse X5S RGB camera captured aerial images at 15 m above ground level (AGL). A ground rover (*Farm-ng, California*) carrying an Insta360 Titan VR 360 8K Camera (*Insta360, Shenzhen*) captured images at 1.5 m AGL while moving through interrows. The visual survey involved identifying groundweeds, with geographic coordinates recorded using an Emlid Reach RS+ Real-Time Kinematic (RTK) positioning global navigation satellite system receiver. In the current deployment pipeline, WeedNet operates in an offline inference mode. We preprocessed UAV images in Pix4D to generate orthomosaics with a ground sampling distance of 0.10–0.12 cm per pixel. We georeferenced each weed and aligned it with corresponding vector data containing spatial location, shape, and canopy boundary information. We then cropped individual weeds from the orthomosaic into single images (204 x 204 in the very early stage and 305 x 305 in the early stage) and processed them through WeedNet for prediction. The WeedNet model was evaluated in two distinct settings: (i) targeted identification of the same 15 specific weeds throughout the evaluation period and (ii) an untargeted (random) visual survey evaluating the presence of various weeds in the field on each survey date. A total of 425 annotated weed images from UAV images and 110 annotated weed images from rover images cover eight species from the early stage in August to the reproductive stage in September. The weed species included: common waterhemp, common lambsquarters, venice mallow (*Hibiscus trionum*), spotted spurge (*Euphorbia maculata*), common purslane (*Portulaca oleracea*), green carpetweed (*Mollugo verticillata*), yellow foxtail, and eastern nightshade (*Solanum emulans*).

To characterize WeedNet's computational performance across different non-embedded computing tiers, we evaluated the model on a CPU workstation, a data-center GPU, and a portable AI workstation. The average inference throughput on CPU (AMD Ryzen 9 7900X 12-Core Processor, 64 GB RAM) was 2.4 img/s, and on the NVIDIA A100 SXM4 80GB GPU device was 201.61 img/s (FP32), 299.8 img/s (FP16), and 412.3 img/s (FP16 + torch.compile), which is sufficient for offline mapping and analysis workflows.

To evaluate deployment beyond datacenter GPUs, we tested the model on an NVIDIA DGX Spark, a compact desktop supercomputer that measures 150 mm × 150 mm × 50.5 mm and weighs 1.2 kg, which integrates a Blackwell GPU with fifth-generation Tensor Cores and a 20-core Arm-based Grace CPU[152], and also reports performance at half-precision (FP16). While ultra-low-power microcontrollers (e.g., NVIDIA Jetson class) represent one class of edge devices, a real-world research platform with an onboard computing system, auxiliary power, and sensors integrated into the vehicle[153] demonstrates that a portable workstation can be carried on a field tractor for field applications. On DGX Spark, the model achieved inference speeds of 62.9 img/s (FP32), 119.8 img/s (FP16), and 162.9 img/s (FP16 + torch. compile), while maintaining 90.91% Top-1 and 98.46% Top-5 accuracy (Supplementary Table S7). Unlike Jetson-class embedded systems, DGX Spark is primarily designed for AI model development, testing, and high-performance local inference. Extending real-time performance to ultra-low-power embedded edge chips (Jetson-class) via pruning and quantization remains an area for future work[154,155].

Integrating the weed identification model into a smartphone app provides a convenient tool for users to connect with the AI model for fast, accurate identification, enabling timely action for weed control. During a five-year period, a weed specialist tested several smartphone apps for plant identification purposes[156]. The dataset included broadleaf, grassy, ornamental, and shrub species. The study emphasized that a simple process is key to user-friendliness and overall experience when using such tools[156]. However, these apps were designed primarily to identify general plant species rather than weed species for agricultural or ecological concerns.

To integrate WeedNet into a smartphone app, we deployed the trained WeedNet model in WeedID (https://go.iastate.edu/COM0Z5) as a REST API using FastAPI, providing a lightweight interface for real-time predictions. The backend processes the input images and returns the predicted species, along with alternative candidates and their corresponding confidence scores. The conformal prediction intervals, described in Section 4.4, quantify uncertainty.

A user-friendly front-end application communicates with this API, enabling users to submit images and receive results (two-step process, Figure 7). Both the Global and Iowa models are publicly available. To ensure system stability, requests are rate-limited to 30 calls per minute. Although the pipeline uses CPU-based inference for cost efficiency, it delivers predictions within practical timeframes (typically under 3 seconds per image). The front and back ends run on OpenShift as isolated microservices, enabling modular maintenance.

The WeedID interface not only identifies weed species but also provides comprehensive plant characteristics relevant to weed management. It presents sixteen attributes that span species taxonomy, ecological traits, identification features, and recommended control approaches (Figure 7). Taxonomic information, including class, order, and family, is sourced from iNaturalist metadata. The Iowa WeedID model includes two additional attributes: local names and a classification based on Iowa's Noxious Weed Law.

PestIDBot generates ecological and identification traits, along with management recommendations, using a large language model-based system that integrates WeedID results with a retrieval-augmented generative chatbot, AgLLM[157,158]. PestIDBot supports this process by communicating relevant management strategies through natural dialogue with users, asking questions about identified and classified weed species, and enabling interactive, informed decision-making. PestIDbot can provide plant information, including life cycle, cotyledon type, leaf type, morphological features, ecological classification, origin in the context of the region, habitat, undesirability, look-alike species, life stage essential for control, and chemical control (Supplementary Note 11). Experts in weed science review this metadata to ensure accuracy. This integrated deployment bridges the gap between weed identification and management strategy, demonstrating how computer vision and LLMs can support weed identification and control for end users.

### 4.7 Model Evaluation

Table 3 summarizes the testing approaches used to evaluate the Global model and its fine-tuned variants. We also evaluated the model's performance on specialized datasets, including intra-species dissimilarity, inter-species similarity, and UAV/Ground Rover images, to assess its robustness in agricultural applications. This study used linear regression to assess the relationship between training image count and accuracy. A paired t-test compared each species' accuracy before and after fine-tuning.

We evaluated model performance using a confusion matrix, which categorized results into four categories: true positives (TP), false positives (FP), true negatives (TN), and false negatives (FN). TP represents the number of correct identifications; FP represents the number of other species misclassified as this species; FN represents the number of these species misclassified as others. We calculated Macro Precision, Macro Recall, and Macro F1 score from the confusion matrices to assess the model's capability and effectiveness. n represents the total number of images.

$$Accuracy = \left(\sum_{i=1}^{n} TP_i\right)/n$$

$$MacroPrecision = \left(\sum_{i=1}^{n} \frac{TP_i}{TP_i + FP_i}\right)/n$$

$$MacroRecall = \left(\sum_{i=1}^{n} \frac{TP_i}{TP_i + FN_i}\right)/n$$

$$MacroF1 = \left(\sum_{i=1}^{n} \frac{2TP_i}{2TP_i + FP_i + FN_i}\right)/n$$

## Data Availability

We sourced the datasets used for training and evaluation in this study from iNaturalist, a publicly available platform (https://www.inaturalist.org). We acknowledge iNaturalist as the data source and its copyright ownership.

## Acknowledgements

We acknowledge iNaturalist as the data source and respect its copyright ownership. This work was supported by the AI-ENGAGE (NSF #2520192), DSFAS Partnership (USDA #2026-67021-45863), AI Institute for Resilient Agriculture (USDA-NIFA #2021-67021-35329), COALESCE: COntext Aware LEarning for Sustainable CybEr-Agricultural Systems (NSF CPS Frontier #1954556), and USDA CRIS Project IOW04714. The R.F. Baker Center for Plant Breeding, Iowa State University.

## Author Contributions Statement

AS and SS designed and conceptualized the project; YS, VNB, ZKD, TTA, MAA, AR, MS and AS acquired the data; TTA, YS, MS, MAA, SC, AB created new software and/or performed analysis; TTA, MS, YS, AR, AS, and SS interpreted data; YS, TTA, VNB, MAA, MS, AR, AS, and SS created the first draft; AS, SS, DM, AK, WE, NM, BG, MA brought out the critical issues relative to the subject, reviewed the draft and revisions, provided novel ideas to improve the work; all authors edited, reviewed, and approved the final draft.

## Competing Interests Statement

The authors declare that there are no competing interests associated with this work.

# References

1. Aldrich, R. J. & Kremer, R. J. *Principles in Weed Management.* 455 (Ames, IA: Iowa State University Press, 1997).
2. Soltani, N. *et al.* Potential corn yield losses from weeds in north america. *Weed Technology* **30**, 979–984 (2016).
3. Flessner, M. L. *et al.* Potential wheat yield loss due to weeds in the united states and canada. *Weed Technology* **35**, 916–923 (2021).
4. Soltani, N. *et al.* Potential yield loss in dry bean crops due to weeds in the united states and canada. *Weed Technology* **32**, 342–346 (2018).
5. National Invasive Species Information Center. What are invasive species? (2025).
6. Green, J. M. & Owen, M. D. Herbicide-resistant crops: Utilities and limitations for herbicide-resistant weed management. *Journal of agricultural and food chemistry* **59**, 5819–5829 (2011).
7. Kniss, A. R. Genetically engineered herbicide-resistant crops and herbicide-resistant weed evolution in the united states. *Weed Science* **66**, 260–273 (2018).
8. Lillie, K., Giacomini, D. & Tranel, P. Comparing responses of sensitive and resistant populations of palmer amaranth (amaranthus palmeri) and waterhemp (amaranthus tuberculatus var. Rudis) to PPO inhibitors. *Weed technology* **34**, 140–146 (2020).
9. Weed Science Society of America. Weed science society of america database.
10. Wang, P., Peteinatos, G., Efthimiadou, A. & Ma, W. Weed identification and integrated control. *Frontiers in Plant Science* vol. 14 1351481 (2023).
11. Roberts, J. & Florentine, S. Advancements and developments in the detection and control of invasive weeds: A global review of the current challenges and future opportunities. *Weed Science* 1–29 (2024).
12. Christensen, S. *et al.* Site-specific weed control technologies. *Weed Research* **49**, 233–241 (2009).
13. Tataridas, A. *et al.* Early detection, herbicide resistance screening, and integrated management of invasive plant species: A review. *Pest Management Science* **78**, 3957–3972 (2022).
14. Adamchuk, V. I., Bernards, M. L., Meyer, G. E. & Mulliken, J. A. *Weed Targeting Herbicide Management*. (2008).
15. WeedScience.org. International herbicide-resistant weed database – resistance case ID 5684. (2025).
16. Livingston, M. *et al. The Economics of Glyphosate Resistance Management in Corn and Soybean Production*. (2015).
17. Bonny, S. Genetically modified herbicide-tolerant crops, weeds, and herbicides: Overview and impact. *Environmental Management* **57**, 31–48 (2016).
18. Jussaume, R. & Dentzman, K. Farmers' perspectives on management options for herbicide-resistant weeds. *Choices* **31**, (2016).
19. Arbuckle, J. G. Farmer perspectives on pesticide resistance. (2014).
20. Robinson, B. L., Moffitt, J. M., Wilkerson, G. G. & Jordan, D. L. Economics and effectiveness of alternative weed scouting methods in peanut. *Weed Technology* **21**, 88–96 (2007).
21. Wu, X., Aravecchia, S., Lottes, P., Stachniss, C. & Pradalier, C. Robotic weed control using automated weed and crop classification. *Journal of Field Robotics* **37**, 322–340 (2020).
22. Zhang, W., Miao, Z., Li, N., He, C. & Sun, T. Review of current robotic approaches for precision weed management. *Current robotics reports* **3**, 139–151 (2022).
23. Franco, C., Pedersen, S. M., Papaharalampos, H. & Ørum, J. E. The value of precision for image-based decision support in weed management. *Precision Agriculture* **18**, 366–382 (2017).

24. Jafar, A., Bibi, N., Naqvi, R. A., Sadeghi-Niaraki, A. & Jeong, D. Revolutionizing agriculture with artificial intelligence: Plant disease detection methods, applications, and their limitations. *Frontiers in Plant Science* **15**, 1356260 (2024).
25. Li, L., Zhang, S. & Wang, B. Plant disease detection and classification by deep learning—a review. *IEEE Access* **9**, 56683–56698 (2021).
26. Bhargava, A. *et al.* Plant leaf disease detection, classification, and diagnosis using computer vision and artificial intelligence: A review. *IEEE access* **12**, 37443–37469 (2024).
27. Toscano-Miranda, R. *et al.* Artificial-intelligence and sensing techniques for the management of insect pests and diseases in cotton: A systematic literature review. *The Journal of Agricultural Science* **160**, 16–31 (2022).
28. Chiranjeevi, S. *et al.* InsectNet: Real-time identification of insects using an end-to-end machine learning pipeline. *PNAS nexus* **4**, pgae575 (2025).
29. Partel, V., Kakarla, S. C. & Ampatzidis, Y. Development and evaluation of a low-cost and smart technology for precision weed management utilizing artificial intelligence. *Computers and electronics in agriculture* **157**, 339–350 (2019).
30. Adhinata, F. D., Sumiharto, R., *et al.* A comprehensive survey on weed and crop classification using machine learning and deep learning. *Artificial intelligence in agriculture* (2024).
31. Murphy, K. M., Ludwig, E., Gutierrez, J. & Gehan, M. A. Deep learning in image-based plant phenotyping. *Annual Review of Plant Biology* **75**, (2024).
32. Deng, B., Lu, Y. & Xu, J. Weed database development: An updated survey of public weed datasets and cross-season weed detection adaptation. *Ecological Informatics* **81**, 102546 (2024).
33. Li, Y. *et al.* Identification of weeds based on hyperspectral imaging and machine learning. *Frontiers in Plant Science* **11**, 611622 (2021).
34. Xu, M., Yoon, S., Wu, C., Baek, J. & Park, D. S. PlantCLEF2023: A bigger training dataset contributes more than advanced pretraining methods for plant identification. in *CLEF (working notes)* 2168–2180 (2023).
35. Goëau, H., Bonnet, P. & Joly, A. Overview of PlantCLEF 2023: Image-based plant identification at global scale. in *CLEF 2023 working notes-24th conference and labs of the evaluation forum* vol. 3497 1972–1981 (2023).
36. Coleman, G. R. *et al.* Weed detection to weed recognition: Reviewing 50 years of research to identify constraints and opportunities for large-scale cropping systems. *Weed Technology* **36**, 741–757 (2022).
37. Lu, Y. & Young, S. A survey of public datasets for computer vision tasks in precision agriculture. *Computers and Electronics in Agriculture* **178**, 105760 (2020).
38. Sunil, G. *et al.* A novel automated cloud-based image datasets for high throughput phenotyping in weed classification. *Data in Brief* **57**, 111097 (2024).
39. Venkataraju, A., Arumugam, D., Kiran, R. & Peters, T. Automated approaches for the early stage distinguishing of palmer amaranth from waterhemp. *Frontiers in Agronomy* **6**, 1425425 (2024).
40. Van Horn, G. *et al.* The inaturalist species classification and detection dataset. in *Proceedings of the IEEE conference on computer vision and pattern recognition* 8769–8778 (2018).
41. Jain, N., Joshi, A. & Earles, M. iNatAg: Multi-class classification models enabled by a large-scale benchmark dataset with 4.7 m images of 2,959 crop and weed species. in *Proceedings of the computer vision and pattern recognition conference* 5501–5510 (2025).
42. Cole, E. *et al.* Spatial implicit neural representations for global-scale species mapping. in *International conference on machine learning* 6320–6342 (PMLR, 2023).
43. Fried, G., Norton, L. R. & Reboud, X. Environmental and management factors determining weed species composition and diversity in france. *Agriculture, ecosystems & environment* **128**, 68–76 (2008).

44. White, E., Soltis, P. S., Soltis, D. E. & Guralnick, R. Quantifying error in occurrence data: Comparing the data quality of iNaturalist and digitized herbarium specimen data in flowering plant families of the southeastern united states. *Plos one* **18**, e0295298 (2023).
45. Gao, M. *et al.* Suppressing label noise in medical image classification using mixup attention and self-supervised learning. *Physics in Medicine & Biology* **69**, 105026 (2024).
46. Hendrycks, D., Mazeika, M., Kadavath, S. & Song, D. Using self-supervised learning can improve model robustness and uncertainty. *Advances in neural information processing systems* **32**, (2019).
47. Ahmed, F., Al-Mamun, H. A., Bari, A. H., Hossain, E. & Kwan, P. Classification of crops and weeds from digital images: A support vector machine approach. *Crop Protection* **40**, 98–104 (2012).
48. Bakhshipour, A. & Jafari, A. Evaluation of support vector machine and artificial neural networks in weed detection using shape features. *Computers and Electronics in Agriculture* **145**, 153–160 (2018).
49. Yano, I. H., Alves, J. R., Santiago, W. E. & Mederos, B. J. Identification of weeds in sugarcane fields through images taken by UAV and random forest classifier. *IFAC-PapersOnLine* **49**, 415–420 (2016).
50. Chen, D., Lu, Y., Li, Z. & Young, S. Performance evaluation of deep transfer learning on multi-class identification of common weed species in cotton production systems. *Computers and Electronics in Agriculture* **198**, 107091 (2022).
51. Hu, K., Coleman, G., Zeng, S., Wang, Z. & Walsh, M. Graph weeds net: A graph-based deep learning method for weed recognition. *Computers and electronics in agriculture* **174**, 105520 (2020).
52. Du, Y., Zhang, G., Tsang, D. & Jawed, M. K. Deep-cnn based robotic multi-class under-canopy weed control in precision farming. in *2022 international conference on robotics and automation (ICRA)* 2273–2279 (IEEE, 2022).
53. Olsen, A. *et al.* DeepWeeds: A multiclass weed species image dataset for deep learning. *Scientific reports* **9**, 2058 (2019).
54. Khan, S. *et al.* A novel semi-supervised framework for UAV based crop/weed classification. *Plos one* **16**, e0251008 (2021).
55. Nong, C., Fan, X. & Wang, J. Semi-supervised learning for weed and crop segmentation using UAV imagery. *Frontiers in Plant Science* **13**, 927368 (2022).
56. Nam, H., Lee, H., Park, J., Yoon, W. & Yoo, D. Reducing domain gap by reducing style bias. *Proceedings of the IEEE/CVF Conference on Computer Vision and Pattern Recognition* 8690–8699 (2021).
57. Vaswani, A. *et al.* Attention is all you need. *Advances in neural information processing systems* **30**, (2017).
58. Khan, S. *et al.* Transformers in vision: A survey. *ACM computing surveys (CSUR)* **54**, 1–41 (2022).
59. Dosovitskiy, A. An image is worth 16x16 words: Transformers for image recognition at scale. *arXiv preprint arXiv:2010.11929* (2020).
60. Li, G., Wang, Y., Zhao, Q., Yuan, P. & Chang, B. PMVT: A lightweight vision transformer for plant disease identification on mobile devices. *Frontiers in Plant Science* **14**, 1256773 (2023).
61. Reedha, R., Dericquebourg, E., Canals, R. & Hafiane, A. Transformer neural network for weed and crop classification of high resolution UAV images. *Remote sensing* **14**, 592 (2022).
62. Huang, X. *et al.* Safety and trustworthiness of deep neural networks: A survey. *arXiv preprint arXiv:1812.08342* 151 (2018).
63. Deng, A., Li, S., Xiong, M., Chen, Z. & Hooi, B. Trust, but verify: Using self-supervised probing to improve trustworthiness. in *European conference on computer vision* 361–377 (Springer, 2022).
64. Nagasubramanian, K. *et al.* How useful is active learning for image-based plant phenotyping? *The Plant Phenome Journal* **4**, e20020 (2021).
65. Borgy, B. *et al.* Changes in functional diversity and intraspecific trait variability of weeds in response to crop sequences and climate. *Weed Research* **56**, 102–113 (2016).

66. Baker, H. G. The evolution of weeds. *Annual review of ecology and systematics* 1–24 (1974).
67. Goolsby, J. A. *et al.* Matching the origin of an invasive weed for selection of a herbivore haplotype for a biological control programme. *Molecular Ecology* **15**, 287–297 (2006).
68. Alvarez-Fernandez, R. *Herbicides: Environmental Impact Studies and Management Approaches*. (BoD–Books on Demand, 2012).
69. Boisseau, R. P., Bradler, S. & Emlen, D. J. Divergence time and environmental similarity predict the strength of morphological convergence in stick and leaf insects. *Proceedings of the National Academy of Sciences* **122**, e2319485121 (2025).
70. Treher, A., Lenny Wilson, R. N. & Kuehn, F. B. Invasive plants and their native look-alikes: An identification guide for the mid-atlantic. (2024).
71. Sun, H. *et al.* Integrative analysis of chloroplast genome, chemicals, and illustrations in bencao literature provides insights into the medicinal value of peucedanum huangshanense. *Frontiers in Plant Science* **14**, 1179915 (2023).
72. University of Maryland Extension. Know your foxtail. (2023).
73. Paleyes, A., Urma, R.-G. & Lawrence, N. D. Challenges in deploying machine learning: A survey of case studies. *ACM computing surveys* **55**, 1–29 (2022).
74. Sarkar, S. *et al.* Cyber-agricultural systems for crop breeding and sustainable production. *Trends in Plant Science* **29**, 130–149 (2024).
75. Caron, M. *et al.* Unsupervised learning of visual features by contrasting cluster assignments. *Advances in neural information processing systems* **33**, 9912–9924 (2020).
76. Weiss, K., Khoshgoftaar, T. M. & Wang, D. A survey of transfer learning. *Journal of Big data* **3**, 1–40 (2016).
77. Sharif Razavian, A., Azizpour, H., Sullivan, J. & Carlsson, S. CNN features off-the-shelf: An astounding baseline for recognition. in *Proceedings of the IEEE conference on computer vision and pattern recognition workshops* 806–813 (2014).
78. Group, P. W. C. & Hub, S. I. Weed-AI: A repository of weed images in crops. (2025).
79. Dang, F., Chen, D., Lu, Y. & Li, Z. YOLOWeeds: A novel benchmark of YOLO object detectors for multi-class weed detection in cotton production systems. *Computers and Electronics in Agriculture* **205**, 107655 (2023).
80. Saadati, M. *et al.* Out-of-distribution detection algorithms for robust insect classification. *Plant Phenomics* **6**, 0170 (2024).
81. Chiranjeevi, S. *et al.* Deep learning powered real-time identification of insects using citizen science data. *arXiv preprint arXiv:2306.02507* (2023).
82. Ilyas, T. *et al.* CWD30: A new benchmark dataset for crop weed recognition in precision agriculture. *Computers and Electronics in Agriculture* **229**, 109737 (2025).
83. Belissent, N., Peña, J. M., Mesías-Ruiz, G. A., Shawe-Taylor, J. & Pérez-Ortiz, M. Transfer and zero-shot learning for scalable weed detection and classification in UAV images. *Knowledge-Based Systems* **292**, 111586 (2024).
84. Saini, P. & Nagesh, D. CottonWeeds: Empowering precision weed management through deep learning and comprehensive dataset. *Crop Protection* **181**, 106675 (2024).
85. Moazzam, S. I., Nawaz, T., Qureshi, W. S., Khan, U. S. & Tiwana, M. I. A w-shaped convolutional network for robust crop and weed classification in agriculture. *Precision Agriculture* **24**, 2002–2018 (2023).
86. Wang, A., Xu, Y., Wei, X. & Cui, B. Semantic segmentation of crop and weed using an encoder-decoder network and image enhancement method under uncontrolled outdoor illumination. *Ieee Access* **8**, 81724–81734 (2020).

87. Umamaheswari, S. & Jain, A. V. Encoder–decoder architecture for crop-weed classification using pixel-wise labelling. in *2020 international conference on artificial intelligence and signal processing (AISP)* 1–6 (IEEE, 2020).
88. Hasan, A. M., Sohel, F., Diepeveen, D., Laga, H. & Jones, M. G. A survey of deep learning techniques for weed detection from images. *Computers and electronics in agriculture* **184**, 106067 (2021).
89. Bi, J. & Zhang, C. An empirical comparison on state-of-the-art multi-class imbalance learning algorithms and a new diversified ensemble learning scheme. *Knowledge-Based Systems* **158**, 81–93 (2018).
90. Huang, C., Li, Y., Loy, C. C. & Tang, X. Learning deep representation for imbalanced classification. in *Proceedings of the IEEE conference on computer vision and pattern recognition* 5375–5384 (2016).
91. Shen, L., Lin, Z. & Huang, Q. Relay backpropagation for effective learning of deep convolutional neural networks. in *Computer vision–ECCV 2016: 14th european conference, amsterdam, the netherlands, october 11–14, 2016, proceedings, part VII 14* 467–482 (Springer, 2016).
92. Hodač, L. *et al.* Exploiting algal strains for robust cross-domain phytoplankton classification via deep learning. *Limnology and Oceanography: Methods* **23**, 815–833 (2025).
93. Joly, A. *et al.* LifeCLEF 2016: Multimedia life species identification challenges. in *International conference of the cross-language evaluation forum for european languages* 286–310 (Springer, 2016).
94. Rzanny, M., Mäder, P., Deggelmann, A., Chen, M. & Wäldchen, J. Flowers, leaves or both? How to obtain suitable images for automated plant identification. *Plant methods* **15**, 77 (2019).
95. Nhan, N. T. T., Le, T.-L. & Hai, V. Do we need multiple organs for plant identification? *2020 International Conference on Multimedia Analysis and Pattern Recognition (MAPR)* 1–6 (2020).
96. Nhan, N. T. T. *et al.* Score-based fusion schemes for plant identification from multi-organ images. *VNU Journal of Science: Computer Science and Communication Engineering* **34**, (2018).
97. Rzanny, M. *et al.* Image-based automated recognition of 31 poaceae species: The most relevant perspectives. *Frontiers in Plant Science* **12**, 804140 (2022).
98. Perez-Ortiz, M. *et al.* Selecting patterns and features for between-and within-crop-row weed mapping using UAV-imagery. *Expert Systems with Applications* **47**, 85–94 (2016).
99. Peteinatos, G. G., Reichel, P., Karouta, J., Andújar, D. & Gerhards, R. Weed identification in maize, sunflower, and potatoes with the aid of convolutional neural networks. *Remote Sensing* **12**, 4185 (2020).
100. Wäldchen, J., Rzanny, M., Seeland, M. & Mäder, P. Automated plant species identification—trends and future directions. *PLoS computational biology* **14**, e1005993 (2018).
101. Chu, G. *et al.* Geo-aware networks for fine-grained recognition. in *Proceedings of the IEEE/CVF international conference on computer vision workshops* (2019).
102. Blank, L., Rozenberg, G. & Gafni, R. Spatial and temporal aspects of weeds distribution within agricultural fields–a review. *Crop Protection* **172**, 106300 (2023).
103. Terry, J. C. D., Roy, H. E. & August, T. A. Thinking like a naturalist: Enhancing computer vision of citizen science images by harnessing contextual data. *Methods in Ecology and Evolution* **11**, 303–315 (2020).
104. Kösters, L. M. *et al.* Data fusion for integrative species identification using deep learning. *Systematic Biology* syaf026 (2025).
105. Oklahoma State University Extension. Herbicide mixing order. (2024).
106. Gutjahr, C., Sökefeld, M. & Gerhards, R. Evaluation of two patch spraying systems in winter wheat and maize. *Weed Research* **52**, 510–519 (2012).
107. Islam, T., Sarker, T. T., Ahmed, K. R., Rankrape, C. B. & Gage, K. Weedvision: Multi-stage growth and classification of weeds using detr and retinanet for precision agriculture. in *2024 international conference on machine learning and applications (ICMLA)* 1374–1381 (IEEE, 2024).

108. Peña, J. M., Torres-Sánchez, J., Serrano-Pérez, A., De Castro, A. I. & López-Granados, F. Quantifying efficacy and limits of unmanned aerial vehicle (UAV) technology for weed seedling detection as affected by sensor resolution. *Sensors* **15**, 5609–5626 (2015).
109. Huang, Y., Reddy, K. N., Fletcher, R. S. & Pennington, D. UAV low-altitude remote sensing for precision weed management. *Weed technology* **32**, 2–6 (2018).
110. Mehta, S. & Rastegari, M. Mobilevit: Light-weight, general-purpose, and mobile-friendly vision transformer. *arXiv preprint arXiv:2110.02178* (2021).
111. Singh, A. K. *et al.* Smart connected farms and networked farmers to improve crop production, sustainability and profitability. *Frontiers in Agronomy* **6**, 1410829 (2024).
112. Qazi, S., Khawaja, B. A. & Farooq, Q. U. IoT-equipped and AI-enabled next generation smart agriculture: A critical review, current challenges and future trends. *Ieee Access* **10**, 21219–21235 (2022).
113. Carion, N. *et al.* End-to-end object detection with transformers. in *European conference on computer vision* 213–229 (Springer, 2020).
114. Seibold, C., Künzel, J., Hilsmann, A. & Eisert, P. From explanations to segmentation: Using explainable AI for image segmentation. *arXiv preprint arXiv:2202.00315* (2022).
115. Xu, P., Zhu, X. & Clifton, D. A. Multimodal learning with transformers: A survey. *IEEE Transactions on Pattern Analysis and Machine Intelligence* **45**, 12113–12132 (2023).
116. Ribeiro, M. T., Singh, S. & Guestrin, C. " why should i trust you?" explaining the predictions of any classifier. in *Proceedings of the 22nd ACM SIGKDD international conference on knowledge discovery and data mining* 1135–1144 (2016).
117. Selvaraju, R. R. *et al.* Grad-cam: Visual explanations from deep networks via gradient-based localization. in *Proceedings of the IEEE international conference on computer vision* 618–626 (2017).
118. iNaturalist. iNaturalist: Observations and user statistics. (2025).
119. National Invasive Species Information Center. Invasive species profiles list. (2024).
120. Midwest Invasive Plant Network. Midwest invasive plant list. (2023).
121. Invasive Species Web Portal. Invasive and exotic weeds. (2023).
122. Iowa State University Extension and Outreach. *Weed Identification Field Guide, 2nd Edition*. (Iowa State University, 2024).
123. Cochran, W. G. & Cochran, W. G. *Sampling Techniques*. vol. 3 (wiley New York, 1977).
124. Agresti, A. *Categorical Data Analysis*. (John Wiley & Sons, 2013).
125. Zheltonozhskii, E., Baskin, C., Mendelson, A., Bronstein, A. M. & Litany, O. Contrast to divide: Self-supervised pre-training for learning with noisy labels. in *2022 IEEE/CVF winter conference on applications of computer vision (WACV)* 387–397 (IEEE, 2022).
126. Singh, M. *et al.* Revisiting weakly supervised pre-training of visual perception models. in *Proceedings of the IEEE/CVF conference on computer vision and pattern recognition* 804–814 (2022).
127. Mahajan, D. *et al.* Exploring the limits of weakly supervised pretraining. in *Proceedings of the european conference on computer vision (ECCV)* 181–196 (2018).
128. Autoencoders, S. D. Learning useful representations in a deep network with a local denoising criterion, pascal vincent, hugo larochelle, isabelle lajoie, yoshua bengio and pierre-antoine manzagol. *J. Mach. Learn. Res. ll* 3371–3408 (2010).
129. Chen, T., Kornblith, S., Norouzi, M. & Hinton, G. A simple framework for contrastive learning of visual representations. in *International conference on machine learning* 1597–1607 (PMLR, 2020).
130. Assran, M. *et al.* Self-supervised learning from images with a joint-embedding predictive architecture. in *Proceedings of the IEEE/CVF conference on computer vision and pattern recognition* 15619–15629 (2023).

131. Zhou, J. *et al.* Ibot: Image bert pre-training with online tokenizer. *arXiv preprint arXiv:2111.07832* (2021).
132. Bao, H., Dong, L., Piao, S. & Wei, F. Beit: Bert pre-training of image transformers. *arXiv preprint arXiv:2106.08254* (2021).
133. He, K. *et al.* Masked autoencoders are scalable vision learners. in *Proceedings of the IEEE/CVF conference on computer vision and pattern recognition* 16000–16009 (2022).
134. Alshammari, H. *et al.* Olive disease classification based on vision transformer and CNN models. *Computational Intelligence and Neuroscience* **2022**, 3998193 (2022).
135. Patel, S. Vision transformer with self-attention for automated plant seedling classification. in *2024 2nd international conference on advancement in computation & computer technologies (InCACCT)* 707–710 (IEEE, 2024).
136. Tong, Z., Song, Y., Wang, J. & Wang, L. Videomae: Masked autoencoders are data-efficient learners for self-supervised video pre-training. *Advances in neural information processing systems* **35**, 10078–10093 (2022).
137. Girdhar, R. *et al.* Omnimae: Single model masked pretraining on images and videos. in *Proceedings of the IEEE/CVF conference on computer vision and pattern recognition* 10406–10417 (2023).
138. Singh, M. *et al.* The effectiveness of mae pre-pretraining for billion-scale pretraining. in *Proceedings of the IEEE/CVF international conference on computer vision* 5484–5494 (2023).
139. Roy, A. G. *et al.* Does your dermatology classifier know what it doesn't know? Detecting the long-tail of unseen conditions. *Medical Image Analysis* **75**, 102274 (2022).
140. Hendrycks, D., Mazeika, M. & Dietterich, T. Deep anomaly detection with outlier exposure. *arXiv preprint arXiv:1812.04606* (2018).
141. Ren, J. *et al.* Likelihood ratios for out-of-distribution detection. *Advances in neural information processing systems* **32**, (2019).
142. Liu, W., Wang, X., Owens, J. & Li, Y. Energy-based out-of-distribution detection. *Advances in Neural Information Processing Systems* **33**, 21464–21475 (2020).
143. Hendrycks, D. & Gimpel, K. A baseline for detecting misclassified and out-of-distribution examples in neural networks. *arXiv preprint arXiv:1610.02136* (2016).
144. Wang, Z. *et al.* Masked face recognition dataset and application. *arXiv preprint arXiv:2003.09093* (2020).
145. Russakovsky, O. *et al.* ImageNet Large Scale Visual Recognition Challenge. *International Journal of Computer Vision (IJCV)* **115**, 211–252 (2015).
146. Van Horn, G. *et al.* Benchmarking representation learning for natural world image collections. in *Proceedings of the IEEE/CVF conference on computer vision and pattern recognition* 12884–12893 (2021).
147. Angelopoulos, A. N., Bates, S., *et al.* Conformal prediction: A gentle introduction. *Foundations and Trends® in Machine Learning* **16**, 494–591 (2023).
148. University of Maryland Extension. Know your foxtails. (2023).
149. University of Maryland Extension. Keys to identifying palmer amaranth and waterhemp. (2023).
150. Sohn, S.-I. *et al.* Identification of amaranthus species using visible-near-infrared (vis-NIR) spectroscopy and machine learning methods. *Remote Sensing* **13**, 4149 (2021).
151. Penn State Extension. Dead nettle, henbit, and ground ivy: Three look-alike weeds.
152. NVIDIA. NVIDIA DGX spark datasheet.
153. Wei, J. *et al.* Towards a viable autonomous driving research platform. in *2013 IEEE intelligent vehicles symposium (IV)* 763–770 (IEEE, 2013).

154. Vadera, S. & Ameen, S. Methods for pruning deep neural networks. *Ieee Access* **10**, 63280–63300 (2022).

155. Wei, L., Ma, Z., Yang, C. & Yao, Q. Advances in the neural network quantization: A comprehensive review. *Applied Sciences* **14**, 7445 (2024).

156. Extension, M. S. U. Plant identification? There's an app for that – actually, several. (2024).

157. PestIDBot: An integrated solution for end-to-end agricultural pest management. (2024).

158. Arshad, M. A. *et al.* AgEval: A benchmark for zero-shot and few-shot plant stress phenotyping with multimodal LLMs. *CoRR* (2024).

159. Giselsson, T. M., Jørgensen, R. N., Jensen, P. K., Dyrmann, M. & Midtiby, H. S. A public image database for benchmark of plant seedling classification algorithms. *arXiv preprint arXiv:1711.05458* (2017).

# Tables

**Table 1.** Comparison of WeedNet classification performance at the genus- and species-levels.

| Prediction Metric | Genus-level | Species-level |
|---|---|---|
| Number | 813 | 1,593 |
| Accuracy | 0.9578 | 0.9102 |
| Macro Precision | 0.9571 | 0.9227 |
| Macro Recall | 0.9540 | 0.9102 |
| Macro F1 | 0.9533 | 0.9066 |

**Table 2.** Results of model performance for weed identification challenges. The model was tested for differences in life stages (intra-species dissimilarity)* and look-alike species (inter-specific similarity)*. Quantitative results include Accuracy, Macro Precision, Macro Recall, and Macro F1.

| Challenges | Class | Number of Testing images | Accuracy | Macro Precision | Macro Recall | Macro F1 |
|---|---|---|---|---|---|---|
| Early plant stage | 84 | 151 | 65.8% | 67.0% | 61.7% | 61.5% |
| Vegetative stage | 84 | 190 | 80.2% | 86.5% | 79.8% | 80.8% |
| Reproductive stage | 84 | 271 | 85.7% | 93.3% | 87.0% | 89.1% |
| Life stage* | 84 | 612 | 79.2% | 93.1% | 79.6% | 83.7% |
| Amaranthus | 8 | 50 | 50.0% | 57.4% | 51.8% | 47.7% |
| Apiaceae | 11 | 61 | 77.1% | 97.4% | 77.4% | 84.9% |
| Grass | 5 | 31 | 45.2% | 59.0% | 44.0% | 50.0% |
| Look-alike* | 24 | 142 | 60.6% | 76.0% | 62.0% | 65.3% |

**Table 3.** Overview of Models Used to Evaluate Performance

| Model Used to Test | Number of Species | Testing Image Dataset |
|---|---|---|
| Global model | 1593 | 20 testing images per species |
| Global-to-Local model | 84 | 20 testing images per species |
| Refined model | 11 | 20 testing images per species |
| Global model | 84 | Web image dataset of intra-species dissimilarity |
| Global model | 24 | Web image dataset of inter-species similarity |
| Global model | 8 | UAV and Ground Rover images |

# Figures

**(a) Intraspecies dissimilarity**

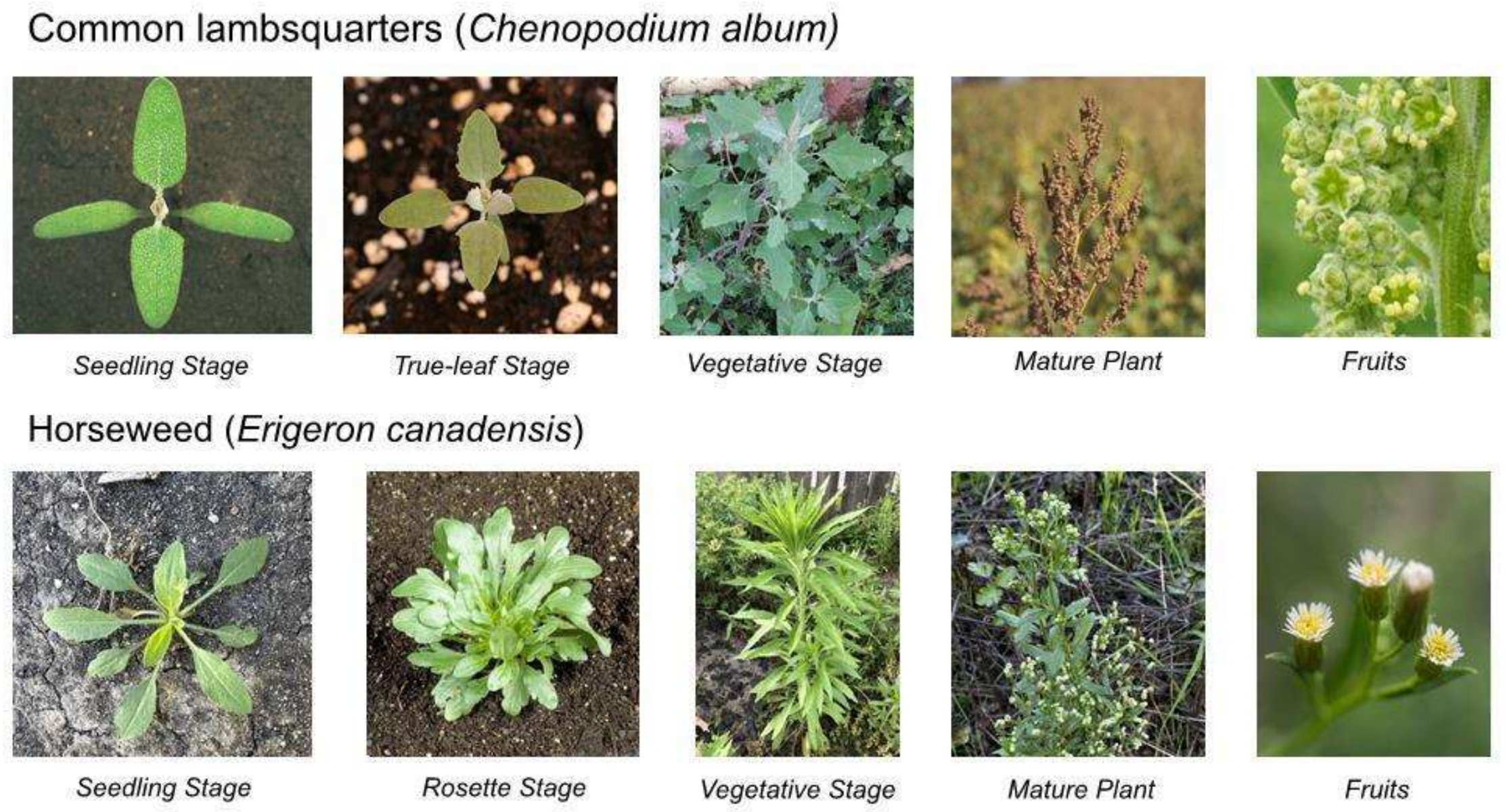

**(b) Interspecies similarity**

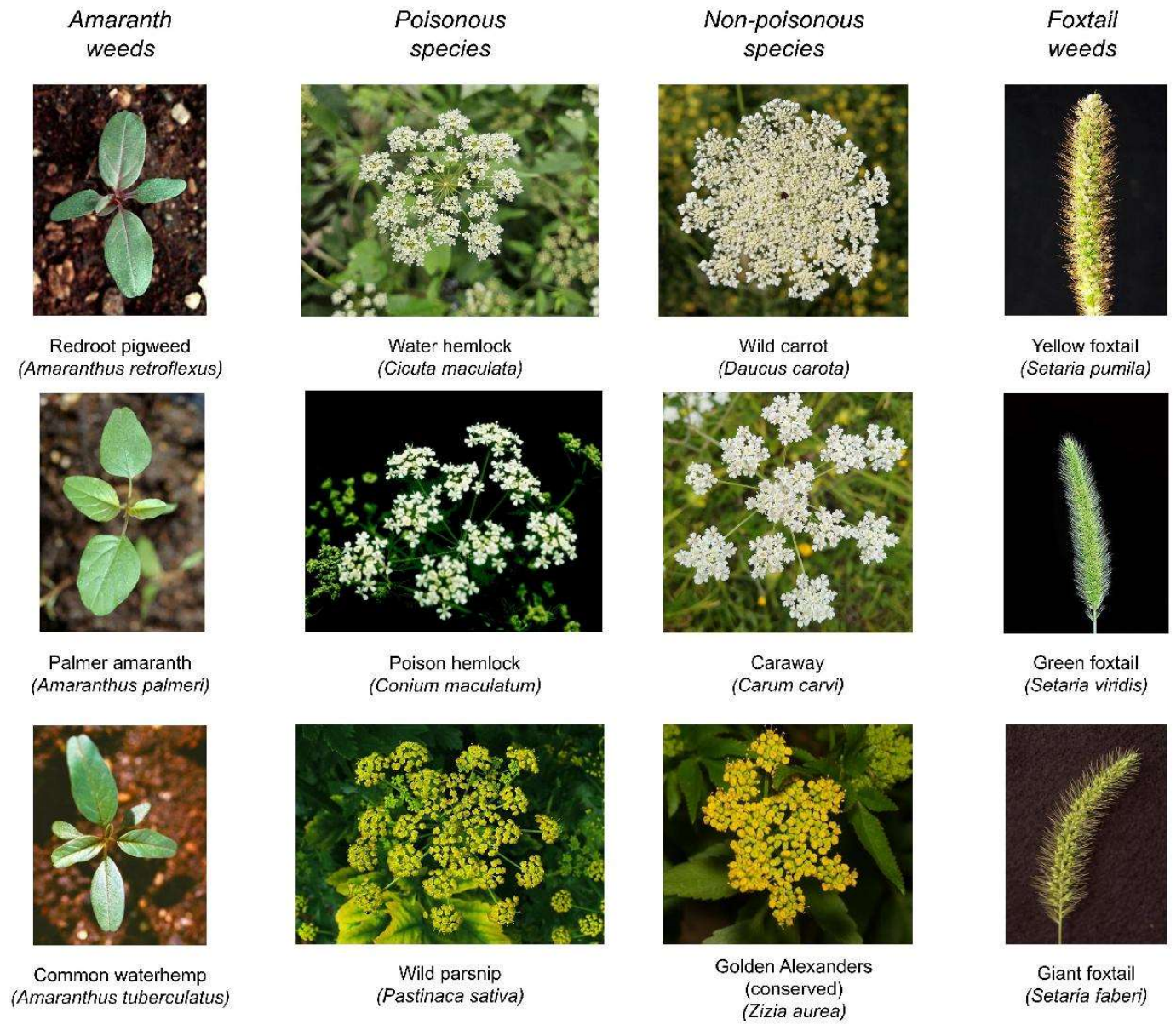

**Figure 1.** Challenging scenarios that arise during weed image classification in the wild. a) Examples of intraspecies dissimilarity across developmental stages in common lambsquarters (*Chenopodium album*) and horseweed (*Erigeron canadensis*). b) Examples illustrating inter-species similarity challenges among species in *Amaranthus* species at the seedling stage, poisonous and non-poisonous species at the inflorescence stage, and foxtail species at the heading stage.

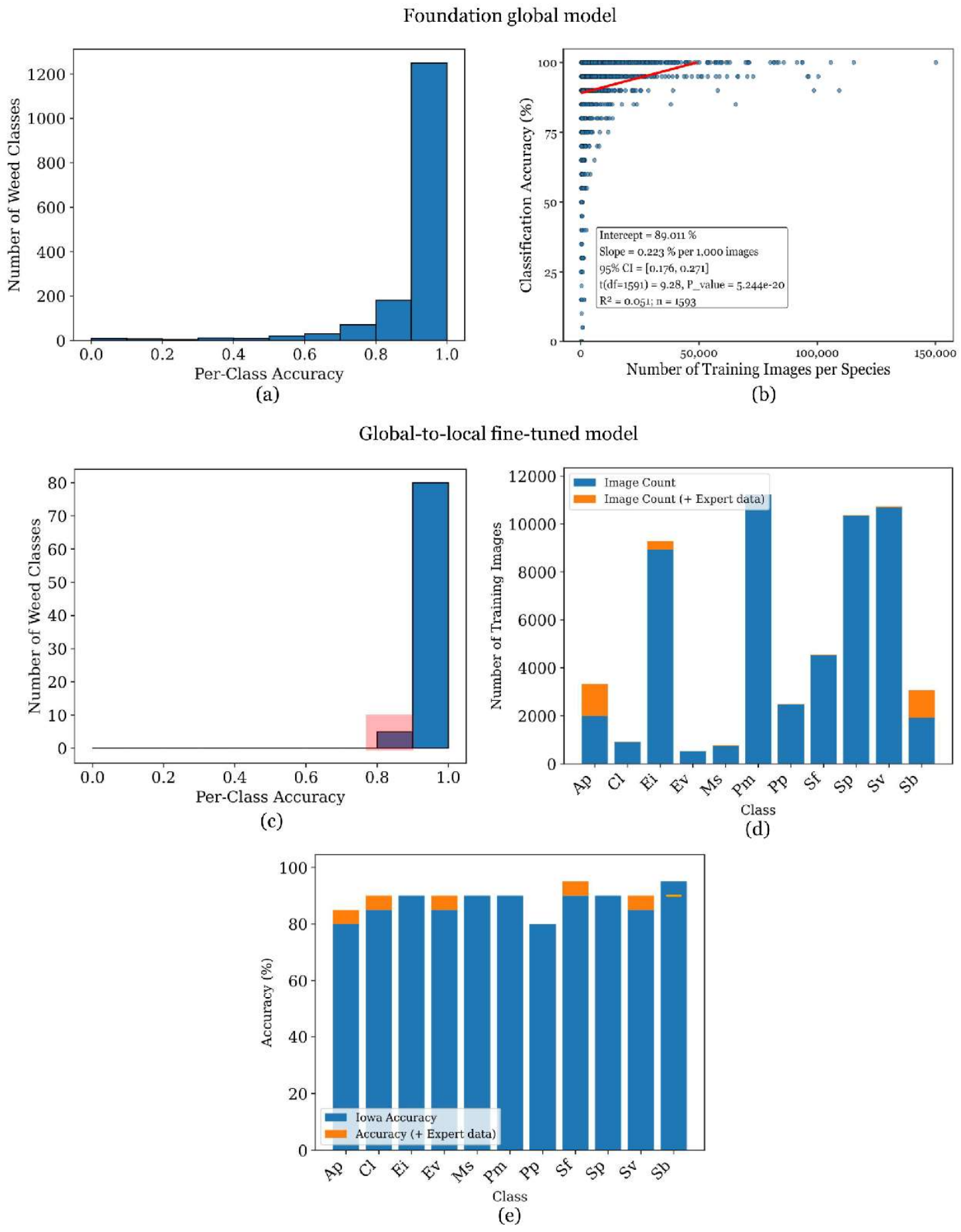

**Figure 2.** Performance analysis across global and Iowa models. (a) Histogram showing the per-class accuracies across 1593 weed species. The plot reveals that nearly 90% of the species achieve over 0.8 (80%) accuracy. (b) Scatter plot illustrating the relationship between prediction accuracy and the amount of training data for the model. Increasing the number of training images does not consistently lead to higher test accuracy. Species with fewer training images achieved 1.0 (100%) accuracy. The figures of the global-to-local fine-tuned model (c) Histogram depicting per-class accuracies across 84 Iowa weed species. (d) Histogram showing the distribution of the number of images in low-performing species, including the count of expert-collected data for these species. (e) Histogram illustrating the accuracy improvement for low-performing species trained with expert data. The 11 species: Ap= *Amaranthus palmeri*, Cl= *Cenchrus longispinus*, Ei= *Eleusine indica*, Ev= *Eriochloa villosa*, Ms= *Muhlenbergia schreberi*, Pm= *Persicaria maculosa*, Pp= *Persicaria pensylvanica*, Sf= *Setaria faberi*, Sp= *Setaria pumila*, Sv= *Setaria viridis*, Sb= *Sorghum bicolor*.

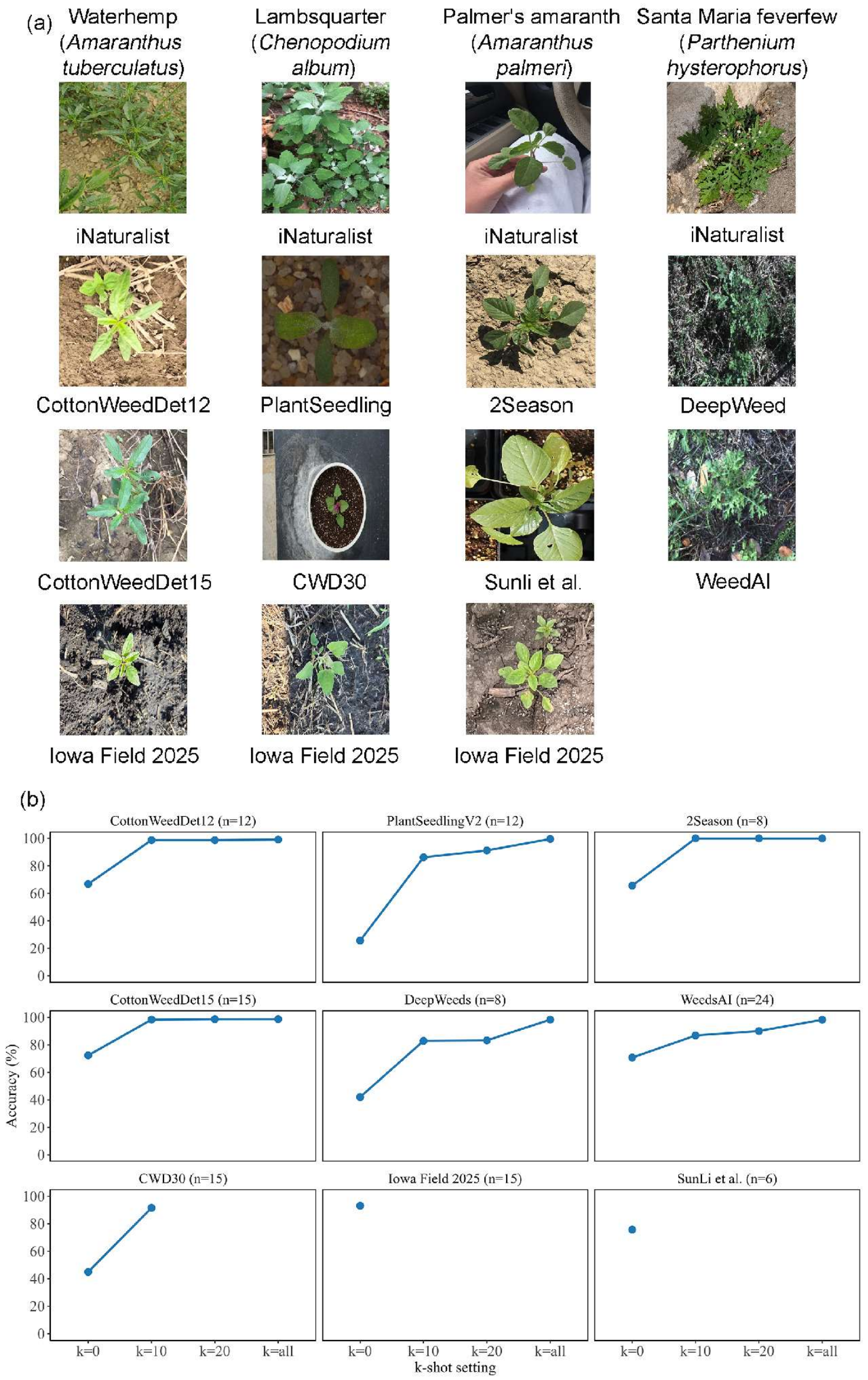

**Figure 3.** Performance of WeedNet on in-field and external weed datasets under k-shot adaptation. (A) Example Weed images of representative weed species from the iNaturalist dataset, compared with multiple agricultural benchmark datasets (CottonWeed12[79], CottonWeed15[50], 2 Seasons[32], DeepWeeds[53], WeedsAI[78], and PlantSeedlingV2[159], CWD30[82], Sunli et al.[38], and Iowa Field 2025), illustrating variation in background, growth stage, and imaging conditions. (B) Top-1 classification accuracy (%) across external datasets under different k-shot settings (k = 0, 10, 20, and all), where k = 0 represents zero-shot evaluation using the foundation model and larger k indicates increasing numbers of target-domain samples used for fine-tuning. Dataset titles report the number of tested species (n). Details are provided in Supplementary Note 12.

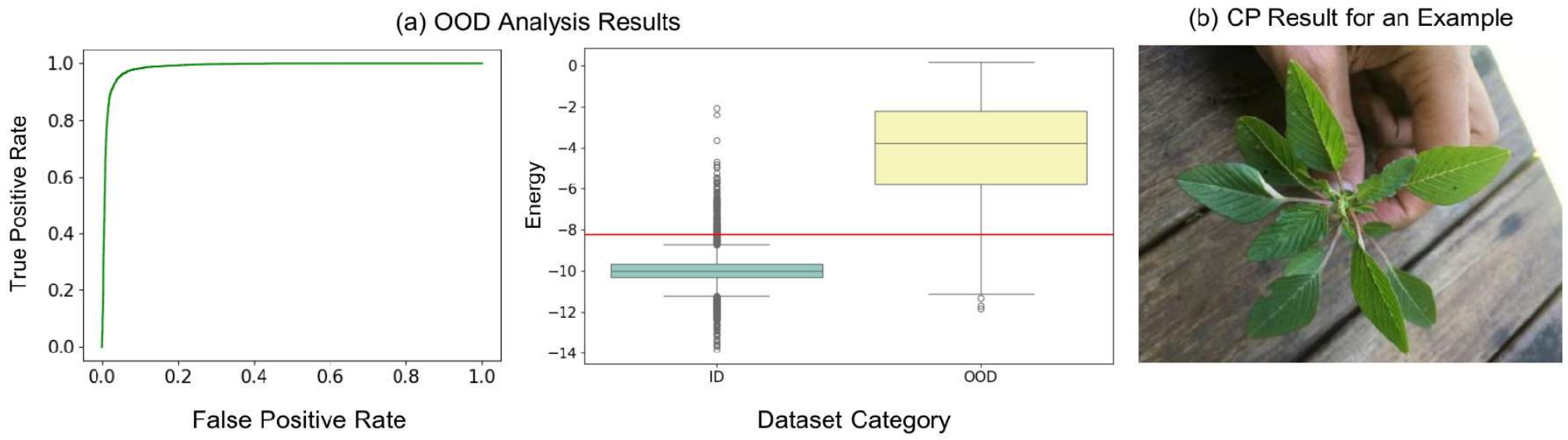

**Figure 4.** ROC curve over all possible threshold values. We chose a threshold value that provides the lowest false positive rate (FPR) at 95% true positive rate (left). Box plot of energy values for in-distribution weed dataset (ID, $n = 3186$ independent image samples) and out-of-distribution (OOD, $n = 3186$ independent image samples) data. Each data point corresponds to the energy score computed from one independent image. No technical replicates were used. In the box plots, the center line represents the median, the box bounds indicate the 25th and 75th percentiles, the whiskers extend to the most extreme values within $1.5 \times$ the interquartile range, and points beyond the whiskers are shown as outliers. The red line indicates the threshold used for separating ID and OOD data (center). Conformal prediction example: *Amaranthus palmeri* (56%), *Amaranthus spinosus* (44%) (right).

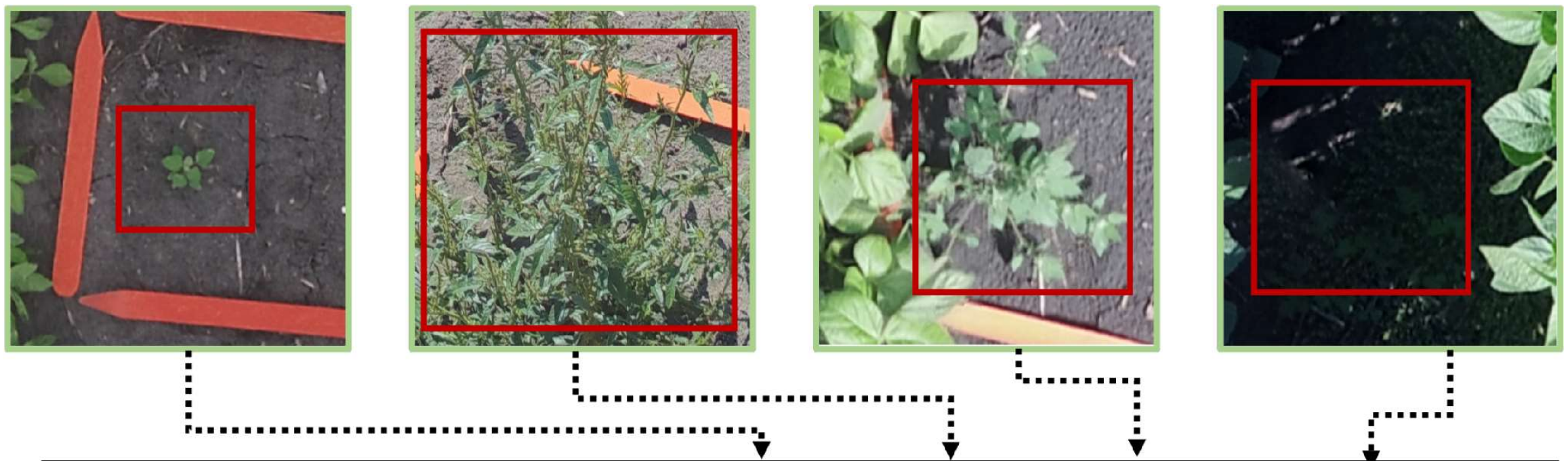

| | | Aug 17 | Aug 24 | Aug 30 | Sep 16 |
|---|---|---|---|---|---|
| **15 Targets** | **Field Survey (Count)** | 15 | 15 | 15 | 15 |
| | **UAV Images (Acc. %)** | 93.3 | | 93.3 | 46.7* |
| | **Ground Robot Images (Acc. %)** | | 82.7 | | |
| **No Target** | **Field Survey (Count)** | 49 | 110 | 100 | 276 |
| | **UAV Images (Acc. %)** | 81.6 | | 69 | 39.1* |
| | **Ground Robot Images (Acc. %)** | | 82.7 | | |
| **Growth Stage** | | Early plant - Vegetation | Early plant - Vegetation | Vegetation - Reproductive | Vegetation - Reproductive |

**Figure 5.** Field evaluation of the WeedNet model using UAV and ground rover data. The model was tested across multiple days (Aug 17, Aug 24, Aug 30, and Sep 16) in the year 2024 under two settings: (i) identification of 15 targeted weed species throughout the evaluation period and (ii) an untargeted visual survey of weeds present in the field. Sample images of randomly observed weeds are shown for each evaluation day. The box within the images highlights the weed in focus. Amiga refers to the Farm-ng Amiga ground-based robot. * indicates that weeds were obscured due to overlapping canopy.

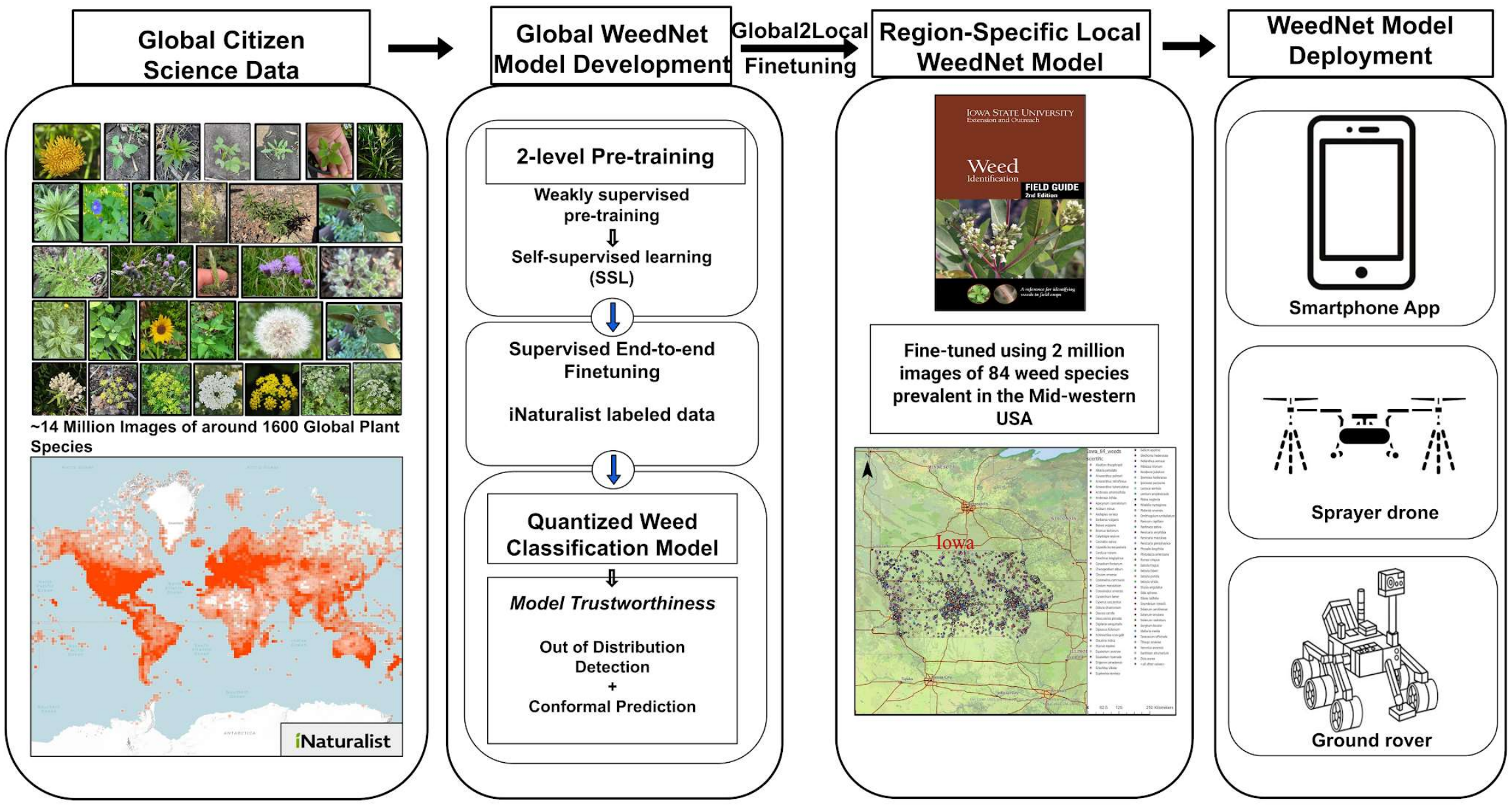

**Figure 6.** Overview of the Global Weed Model Development and Implementation. The Global Weed AI Model is trained on approximately 14 million images covering around 1600 plant species from the iNaturalist database. The end-to-end WeedNet pipeline was deployed on smartphones for real-time weed identification. The Global-to-Local approach enables fine-tuning on region-specific weed species using fewer images, reducing computational requirements while improving accuracy. Such local, region-specific lightweight models could identify weed images captured using smartphones and robotic platforms, such as UAVs and ground rovers, potentially deploying these models for high-throughput, real-time weed identification.

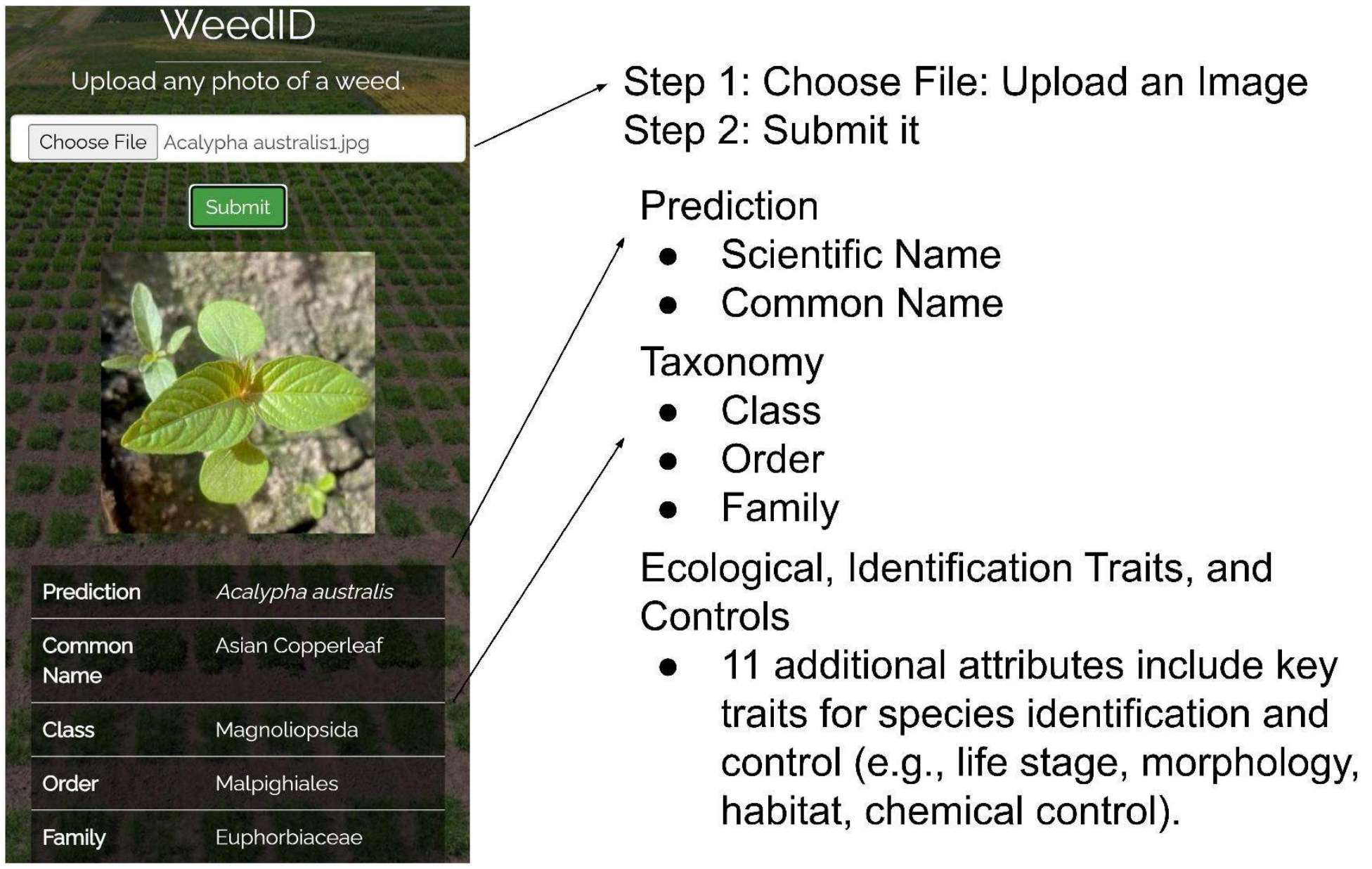

**Figure 7.** Example interface of the WeedID application on an Asian copperleaf image. Users upload weed images for identification and return a prediction. The app provides the species names, species taxonomy, ecological traits, morphology, and control information. The global model is publicly available at https://go.iastate.edu/COM0Z5, and the Iowa fine-tuned model can be accessed at https://go.iastate.edu/O1NZ2Q.